\documentclass[sigconf]{acmart}
\AtBeginDocument{%
	}

\usepackage{amsfonts}
\usepackage{amsmath}

\usepackage[T1]{fontenc}  
\usepackage[utf8]{inputenc}  

\usepackage{graphicx}  
\usepackage{svg}  
\svgsetup{inkscapelatex=true}

\usepackage{algorithm}
\usepackage{algpseudocode}

\usepackage[table,dvipsnames]{xcolor}  
\usepackage{threeparttable}
\usepackage{booktabs}  

\usepackage{subcaption}

\usepackage{mdframed}
\usepackage[most]{tcolorbox}
\tcbuselibrary{listings,breakable}

\usepackage{xurl}

\newtcblisting{promptbox}[1][]{
	listing only,
	breakable,
	colback=gray!3,
	colframe=gray!60,
	boxrule=0.3pt,
	arc=2mm,
	outer arc=2mm,
	left=1mm,
	right=1mm,
	top=1mm,
	bottom=1mm,
	listing options={
		basicstyle=\ttfamily\footnotesize,
		breaklines=true,
		breakatwhitespace=false
	},
	fonttitle=\bfseries\footnotesize,
	title={#1},
}

\renewcommand\footnotetextcopyrightpermission[1]{}

\title{TikArt: Stabilizing Aperture-Guided Fine-Grained Visual Reasoning with Reinforcement Learning}

\author{Hao Ding}
\authornotemark[1] 
\affiliation{%
	\institution{Zhejiang University}
	\city{Hangzhou}
	\country{China}
}
\email{dinghao24@zju.edu.cn}

\author{Zhichuan Yang}
\authornotemark[1]
\affiliation{%
	\institution{Xi'an Jiaotong University}
	\city{Xi'an}
	\country{China}
}

\author{Weijie Ge}
\affiliation{%
	\institution{Zhejiang University of Science and Technology}
	\city{Hangzhou}
	\country{China}
}

\author{Ziqin Gao}
\affiliation{%
	\institution{Zhejiang University}
	\city{Hangzhou}
	\country{China}
}

\author{Chaoyi Lu}
\affiliation{%
	\institution{Xi'an Jiaotong University}
	\city{Xi'an}
	\country{China}
}

\author{Lei Zhao}
\authornotemark[2]
\affiliation{%
	\institution{Zhejiang University}
	\city{Hangzhou}
	\country{China}
}
\email{cszhl@zju.edu.cn}

\thanks{* Equal contribution: Hao Ding and Zhichuan Yang. \\
	$\dagger$ Corresponding author: Lei Zhao (lei.zhao@zju.edu.cn). \\
     The source code is available at \href{https://github.com/TikArt-Team/TikArt}{\texttt{TikArt-Team/TikArt}}.
}

\begin{abstract}
Fine-grained visual reasoning in multimodal large language models (MLLMs) is bottlenecked by single-pass global image encoding: key evidence often lies in tiny objects, cluttered regions, subtle markings, or dense charts. We present \textbf{TikArt} (\textbf{T}h\textbf{i}n\textbf{k}ing \textbf{A}pe\textbf{rt}ure), an aperture-guided agent that formulates multimodal reasoning as sequential evidence acquisition over regions of interest. TikArt follows a Think--Aperture--Observe (TAO) loop that interleaves language reasoning with two aperture actions: Zoom, which extracts rectangular crops, and Segment, which invokes an off-the-shelf segmenter to produce object-centric mask-based views for irregular targets. A mandatory Observation step after every aperture action writes local evidence back into text, yielding interpretable aperture trajectories and persistent linguistic memory.

Built on Qwen3-VL-8B, TikArt is trained with GRPO-style reinforcement learning under a two-stage curriculum. To stabilize long-horizon tool-integrated learning, we introduce Relative Uncertainty Reduction (RUR), a dense reward computed by a frozen evaluator that favors evidence-building trajectories and mitigates degenerate tool use. Experiments on high-resolution reasoning, general multimodal understanding, and both referring and reasoning-oriented segmentation show consistent gains over the backbone, demonstrating that aperture-guided observation improves fine-grained visual reasoning and transfers naturally to pixel-level grounding.

\end{abstract}


\begin{document}
\begin{teaserfigure}
  \centering
  \includegraphics[width=\textwidth]{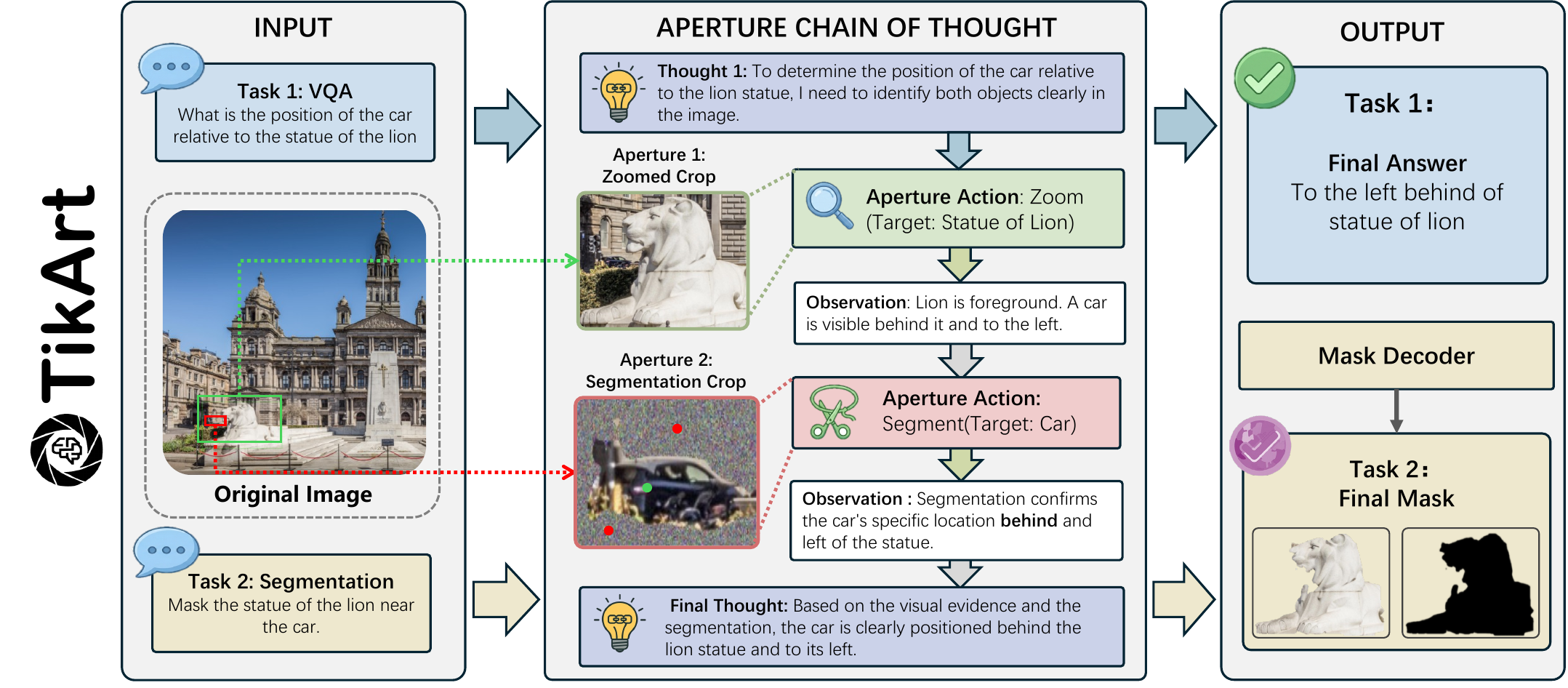}
  \caption{TikArt alternates between aperture selection and grounded observation across tasks. In this example, one trajectory supports both spatial VQA (blue) and segmentation (yellow). Dashed links indicate RoIs extracted from the same high-resolution image, and SAM2~\citep{SAM-2} provides the mask-based crop.}
  \label{fig:case_shower}
\end{teaserfigure}

\maketitle

\section{Introduction}

Multimodal large language models (MLLMs), such as GPT-4o~\cite{openai2024gpt4osystemcard}, QwenVL~\cite{bai2023qwen}, Gemini~\cite{gemini-2.5}, and InternVL~\cite{internVL3.5}, have made substantial progress on general vision--language tasks. Yet \emph{fine-grained} visual reasoning remains a persistent bottleneck: decisive evidence is often localized in tiny objects, subtle markings, dense charts, or cluttered regions. Most MLLMs encode the full image once into a fixed set of visual tokens and then perform most reasoning in the text domain, which makes it difficult to reliably re-inspect critical details. High-resolution benchmarks such as V$^{*}$~\cite{wu2024v} and HR-Bench~\cite{wang2025divide} show that scaling model size or context length alone does not resolve this problem.

Humans instead solve such tasks by actively deciding \emph{where to look}, inspecting local evidence, and using the new evidence to guide the next step. Motivated by this process, we introduce \textbf{TikArt} (Thinking Aperture), an aperture-guided agent that casts multimodal reasoning as sequential evidence acquisition over regions of interest (RoIs). TikArt follows a Think--Aperture--Observe (TAO) loop that alternates between language reasoning, localized perception, and grounded observation writing (Fig.~\ref{fig:case_shower}, Fig.~\ref{fig:TAOLoop}).

\begin{figure}[htbp]
  \includesvg[width=\columnwidth]{figures/TAOLoop.svg}
  \caption{TikArt follows a Think--Aperture--Observe (TAO) loop. Beyond box-based Zoom, TikArt includes a mask-based Segment action to obtain object-centric apertures for irregular or cluttered targets, and explicitly writes observations back into the context after each action.}
  \label{fig:TAOLoop}
\end{figure}

A key limitation of prior ``zoom-only'' pipelines is that rectangular crops are often insufficient for irregular, thin, occluded, or heavily cluttered targets. To address this, TikArt equips the policy with two complementary aperture actions. \textbf{Zoom} extracts rectangular crops for structured evidence such as charts, panels, and table cells. \textbf{Segment} invokes an off-the-shelf segmentation model (SAM2~\citep{SAM-2}) to obtain object-centric mask-based views that better isolate irregular instances. Importantly, Segment is treated as a general perception action rather than a task-specific output: it can support both VQA-style reasoning and segmentation benchmarks.

TikArt further imposes a \textbf{mandatory Observation} step after every aperture action. Rather than leaving newly acquired evidence latent in the hidden state, the model must explicitly describe what appears inside the selected view before continuing. This turns local perception into persistent textual memory, produces interpretable \emph{Aperture Chain-of-Thought} (A-CoT) trajectories, and tightens the coupling between visual actions and downstream reasoning.

Training such long-horizon, tool-integrated policies is non-trivial. Sparse end rewards provide weak credit assignment, and GRPO-style optimization can degenerate when rewards collapse within a rollout group. We therefore introduce \textbf{Relative Uncertainty Reduction (RUR)} (Sec.~\ref{sec:rur}), a dense reward computed by a frozen evaluator that measures whether the trajectory prefix increases confidence in the intended task target. RUR supplies a stationary, trajectory-sensitive signal that favors evidence-building rollouts and stabilizes tool use during RL.

We instantiate TikArt on Qwen3-VL-8B~\cite{bai2025qwen3vltechnicalreport} and optimize it with a two-stage curriculum: a segmentation warm-up stage followed by multi-task GRPO over visual math, fine-grained VQA, and segmentation. We also include a prompt-only inference baseline (\textit{w/o GRPO}) to isolate the contribution of the TAO interface from RL.

Experiments show that TikArt substantially improves fine-grained reasoning. On V$^{*}$ and HR-Bench 4K/8K, TikArt-8B consistently outperforms Qwen3-VL-8B-Instruct, and the learned RoI-centric policy also transfers to pixel-level grounding on RefCOCO and ReasonSeg.

\paragraph{Contributions.}
Our contributions are:
\begin{enumerate}
    \item We introduce a \textbf{dual-aperture action space} for fine-grained multimodal reasoning, combining \textbf{Zoom} for structured local inspection and \textbf{Segment} for mask-based, object-centric evidence acquisition on irregular or cluttered targets.
    \item We propose the \textbf{mandatory Observation contract} and the resulting \textbf{Aperture Chain-of-Thought (A-CoT)} interface, which writes local visual evidence into explicit, auditable textual memory and tightens credit assignment in long-horizon reasoning.
    \item We develop \textbf{TikArt}, an aperture-guided agent trained with GRPO without chain-of-thought supervision, and introduce \textbf{Relative Uncertainty Reduction (RUR)} as a dense trajectory-validity reward for stabilizing tool-integrated rollouts across both discrete-answer reasoning and segmentation settings.
    \item Across high-resolution reasoning, real-world multimodal understanding, and reasoning-oriented segmentation, we show that the learned aperture policy improves fine-grained evidence acquisition and transfers naturally from question answering to pixel-level grounding.
\end{enumerate}

\section{Related Work}

\paragraph{Iterative multimodal reasoning with localized perception.}
A growing line of multimodal reasoning systems augments static image understanding with iterative visual operations or external tools.
ReAct-style reasoning--action interleaving has inspired multimodal extensions such as MM-ReAct~\cite{mmReact1}, while more recent ``think-with-image'' methods incorporate localized crop/zoom operations into the reasoning loop~\cite{skecth,thinkingimagesmultimodalreasoning,zoomrf1,zheng_deepeyes_2025,hong_deepeyesv2_2025}.
These works show the value of query-conditioned perception, but most focus on box-centric inspection and do not require the acquired local evidence to be written back as explicit grounded observations.
TikArt differs by combining box- and mask-centric apertures with a mandatory Observation contract, making intermediate evidence both more expressive and more auditable.

\paragraph{Segmentation models and segmentation-capable MLLMs.}
Foundation segmentation models such as SAM~\cite{kirillov2023segment} and SAM2~\citep{SAM-2} provide strong promptable masks across diverse domains.
LISA~\cite{lai2024lisa} connects segmentation with MLLMs for end-to-end reasoning segmentation, and Seg-R1 learns spatial prompts that guide SAM2 for segmentation-oriented tasks~\cite{Seg-R1}.
Unlike prior systems that mainly treat segmentation as a task output, TikArt treats Segment as a general perception action that can support both VQA-style evidence acquisition and pixel-level grounding.

\paragraph{Process supervision and RL stability for tool-using agents.}
Long-horizon tool-integrated rollouts often suffer from weak credit assignment under sparse outcome rewards.
Process Reward Models and related agentic evaluators provide denser supervision for intermediate steps and tool interactions~\cite{choudhury2025agentprm,liu2025istar,yao2026prl}.
Recent work has also explored dense rewards for fine-grained visual reasoning, such as RewardMap~\cite{feng2026rewardmap}.
TikArt follows this direction but does not train a separate PRM; instead, it combines explicit Observation writing with Relative Uncertainty Reduction (RUR), a frozen-evaluator reward that measures whether the collected trajectory prefix increases confidence in the intended task target.

\paragraph{RL for fine-grained visual reasoning and visual grounding.}
Recent work applies reinforcement learning not only to improve final-answer accuracy, but also to guide visual inspection.
RewardMap~\cite{feng2026rewardmap}, DIP-R1~\cite{park2025dipr1}, and ViGoRL~\cite{sarch2025grounded} all show the value of RL for fine-grained visual reasoning and grounded exploration.
TikArt is closely related to these efforts, but differs in coupling visual search with explicit Observation writing and in exposing dual aperture primitives for controllable evidence acquisition.

\paragraph{Fine-grained perception and recognition in MLLMs.}
Recent MLLM research has also targeted fine-grained perception more directly.
Fine-R1 improves fine-grained visual recognition through chain-of-thought supervised fine-tuning and triplet-augmented policy optimization~\cite{he2026finer1}, while GFMLLM proposes a unified global-and-fine-grained spatial perception framework without expert tools~\cite{fan2026gfmllm}.
Compared with these approaches, TikArt focuses on tool-augmented evidence acquisition by making intermediate observations mandatory and leveraging external aperture and segmentation tools before answering.

\section{Method}
\label{sec:method}

\begin{figure*}[htbp]
  \centering
  \includegraphics[width=\linewidth]{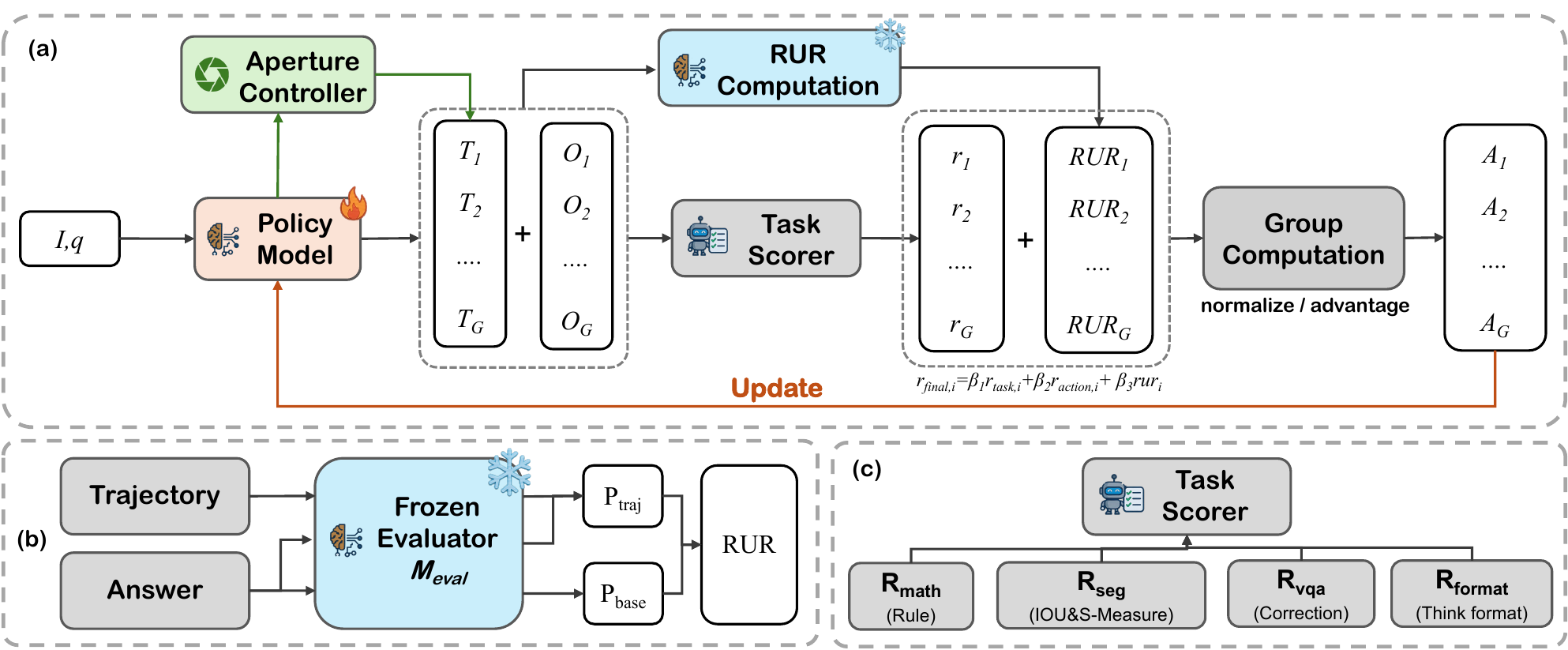}
  \caption{\textbf{RUR-augmented GRPO training overview.}
    (a) We generate a group of $G$ rollouts with the current multimodal policy and compute group-normalized advantages for GRPO updates.
    (b) RUR is computed by a separate \emph{frozen} evaluator via teacher forcing on the task target $y^{*}$, comparing the baseline context (input only) against the trajectory context (trajectory prefix; final answer tokens excluded to prevent leakage).
    (c) The task scorer can combine heterogeneous supervision sources (e.g., rule-based math, segmentation metrics, VQA judging, and format constraints) into task/action rewards, which are then aggregated with RUR using fixed weights.}
  \label{fig:training}
\end{figure*}

TikArt equips a vision-language model (Qwen3-VL-8B~\cite{bai2025qwen3vltechnicalreport}) with
\emph{aperture-guided} visual interaction. Instead of relying on a single, static image encoding,
TikArt interleaves language reasoning with explicit \emph{aperture actions} that extract local visual
evidence and write it back into the context. This produces a unified \emph{Aperture Chain-of-Thought}
(A-CoT), where the model repeatedly decides \emph{whether} to look, \emph{where} to look, and \emph{what}
it sees, before emitting the final answer.

\subsection{A-CoT as a Markov Decision Process}

Given an image $I$ and a query $q$, TikArt generates a token sequence from a combined vocabulary
$\mathcal{V}\cup\mathcal{T}$, where $\mathcal{V}$ is the text-token vocabulary and $\mathcal{T}$ contains
decision tokens such as \texttt{<ZOOM>}, \texttt{<SEGMENT>}, and \texttt{<ANSWER>}. Each generation step
corresponds to an MDP transition.

\paragraph{State.}
At time $t$, the multimodal state is
\begin{equation}
s_t = (I, q, h_{t-1}, \mathcal{A}_{1:t-1}),
\label{eq:state}
\end{equation}
where $h_{t-1}$ is the partially generated A-CoT text, and $\mathcal{A}_{1:t-1}$ denotes all previously
extracted apertures (regions-of-interest and their derived views / descriptions).

\paragraph{Action and transition.}
The action $a_t\in \mathcal{V}\cup\mathcal{T}$ is the next token sampled from the policy
$\pi_\theta(a_t\mid s_t)$. If $a_t$ triggers an aperture (\texttt{<ZOOM>} or \texttt{<SEGMENT>}),
the environment returns a local view $v_t$; otherwise $v_t=\varnothing$.
A rollout is
\begin{equation}
\tau=\{(s_t, a_t, v_t)\}_{t=1}^{T},
\label{eq:trajectory}
\end{equation}
terminated when \texttt{<ANSWER>} is produced.

\paragraph{Structured reasoning loop.}
TikArt follows a recurrent interaction pattern:
\[
\textbf{Think} \rightarrow \textbf{Aperture} \rightarrow \textbf{Observe} \rightarrow \cdots \rightarrow \textbf{Answer},
\]
enabling dynamic state updates driven by targeted visual exploration.

\subsection{Visual Apertures}
\label{sec:aperture_actions}

Apertures are \emph{explicit perception actions} that control where the model looks and what visual evidence is exposed to the policy.
Rather than relying on a single-pass global image encoding, TikArt treats localized evidence acquisition as part of the action space: after selecting an aperture, the environment returns a local view that is subsequently written back into the context through Observation (Sec.~\ref{sec:observation_contract}).

\paragraph{Why box-only zoom is insufficient.}
Rectangular crops are effective for structured evidence such as table cells, text blocks, chart components, or interface panels.
However, many fine-grained visual targets are irregular, thin, partially occluded, or embedded in clutter.
In such cases, a box crop often still contains substantial distractors and fails to isolate the true evidence needed for reliable reasoning.
This limitation becomes especially severe when the model must distinguish one instance from many similar candidates, or when the downstream task itself requires object-level grounding.
Motivated by this, TikArt expands the action space beyond zoom-only inspection.

\paragraph{Zoom (box-centric aperture).}
For structured regions, TikArt predicts a bounding box
\begin{equation}
b_t=(x_{1},y_{1},x_{2},y_{2})\in[0,1000]^4,
\label{eq:zoom}
\end{equation}
which defines the crop
\begin{equation}
v_t^{\text{zoom}}=\mathrm{Crop}(I,b_t).
\end{equation}
The $[0,1000]$ parameterization normalizes coordinates across input resolutions.

\paragraph{Segment (mask-centric, object-centric aperture via SAM2).}
For irregular objects or cluttered scenes, TikArt triggers a \textbf{Segment} action and predicts prompts
$(b_t,p_t,\ell_t)$ for SAM2~\citep{SAM-2}, where $b_t$ is a coarse box, $p_t$ are point prompts, and $\ell_t$ are binary point labels.
SAM2 returns a binary mask $M_t \in \{0,1\}^{H\times W}$.
We then construct an object-centric aperture view by suppressing background while preserving the foreground appearance:
\begin{equation}
v_t^{\text{seg}}=M_t\odot I + (1-M_t)\odot N,
\label{eq:seg}
\end{equation}
where $N$ is Gaussian noise and $\odot$ denotes element-wise multiplication.
This design preserves the visual evidence of the target while reducing distractors, making the subsequent Observation step more grounded and less prone to hallucination.
For segmentation tasks, $M_t$ is also used directly as the predicted mask for evaluation.

\paragraph{Complementarity and cross-task reuse.}
Zoom and Segment are complementary rather than redundant: Zoom is efficient for structured evidence, while Segment is more effective for irregular, thin, occluded, or cluttered targets.
In implementation, both actions are emitted in a structured slot format with quantized geometry, and invalid predictions (e.g., out-of-range or degenerate prompts) are rejected before tool execution.
Importantly, Segment is not only a task-specific output but a general perception action: the same mask-centric aperture can support VQA-style reasoning by isolating the relevant instance, and segmentation tasks by producing the final mask.

\subsection{Mandatory Observation: Grounding What the Model Sees}
\label{sec:observation_contract}

A defining constraint of TikArt is the \textbf{mandatory Observation contract} after every aperture action.
Once a local view $v_t$ is obtained (either $v_t^{\text{zoom}}$ or $v_t^{\text{seg}}$), the model must emit an observation text segment
$o_t$ that describes the content inside the aperture \emph{before it is allowed to take the next action or answer}.

\paragraph{Observation as a constrained decoding rule (a contract, not a narration).}
Formally, if $a_t\in\{\texttt{<ZOOM>},\texttt{<SEGMENT>}\}$, then the subsequent tokens must form $o_t$ until a delimiter (e.g., \texttt{<OBS\_END>}).
At execution time, decoding switches to an observation mode in which further action tokens and \texttt{<ANSWER>} are masked until \texttt{<OBS\_END>} is produced or a maximum observation length is reached; malformed traces receive the format penalty in Sec.~\ref{sec:reward_design}.
The observation is appended to the trajectory context to form the next state:
\begin{equation}
h_t \;=\; h_{t-1}\;\oplus\; a_t \;\oplus\; o_t,
\qquad
s_{t+1} \;=\; (I,q,h_t,\mathcal{A}_{1:t}),
\label{eq:observe_update}
\end{equation}
where $\oplus$ denotes concatenation and $\mathcal{A}_{1:t}$ is updated with the new aperture view.
Therefore, Observation is not merely explanatory text; it is the interface by which newly acquired local evidence becomes persistent textual memory.

\paragraph{Execution-time enforcement.}
The Observation contract is implemented as a constrained decoder state machine rather than a prompt-only preference.
After \texttt{<ZOOM>} or \texttt{<SEGMENT>}, decoding switches to an observation mode in which further action tokens and \texttt{<ANSWER>} are masked until \texttt{<OBS\_END>} is produced or a maximum observation length $L_{\text{obs}}$ is reached.
Malformed traces that omit \texttt{<OBS\_END>} are terminated and assigned the format penalty described in Sec.~\ref{sec:reward_design}.
This mechanism ensures that every successful aperture call contributes an explicit textual evidence span to the trajectory.

\paragraph{Why the Observation contract is central.}
Mandatory Observation directly changes the learning problem in three ways.
First, it makes evidence \emph{persistent and auditable}: the model cannot hide useful visual evidence in latent states, because each aperture must be followed by an explicit grounded description.
Second, it tightens \emph{credit assignment}: a poorly chosen aperture typically produces uninformative or inconsistent observations, making downstream failure attributable to the action itself.
Third, it makes trajectory-level reward shaping meaningful: RUR evaluates whether the collected trajectory prefix improves a frozen evaluator's confidence in the intended task target, which requires intermediate evidence to be explicitly written into the context.

\paragraph{Observation and training stability.}
Beyond interpretability, the Observation contract also stabilizes long-horizon training.
Without explicit observations, aperture calls are easier to degenerate into weakly grounded or noisy tool usage, because their contribution is not forced into the trajectory.
By contrast, mandatory Observation couples each action to a textual evidence trace, which reduces uncontrolled aperture wandering and improves the learnability of the policy.
This interpretation is consistent with the training dynamics in Fig.~\ref{fig:training_dynamics}, where removing Observation leads to higher entropy, heavier-tailed aperture usage, and weaker rewards.

\paragraph{TAO loop with mandatory Observation.}
Algorithm~\ref{alg:tao} summarizes the Think--Aperture--Observe interaction pattern enforced in TikArt.

\begin{algorithm}[t]
\caption{Think--Aperture--Observe (TAO) with the mandatory Observation contract.}
\label{alg:tao}
\begin{algorithmic}[1]
\Require Image $I$, query $q$, max interaction steps $T$
\State Initialize trajectory text $h_0$ with task prompt and formatting constraints
\For{$t=1$ to $T$}
    \State Generate \texttt{<think>} tokens until producing an action token or \texttt{<ANSWER>}
    \If{action token is \texttt{<ZOOM>} with box $b_t$}
        \State $v_t \gets \mathrm{Crop}(I, b_t)$
    \ElsIf{action token is \texttt{<SEGMENT>} with prompts $(b_t,p_t,\ell_t)$}
        \State $M_t \gets \mathrm{SAM2}(I, b_t, p_t, \ell_t)$
        \State $v_t \gets \mathrm{MaskView}(I, M_t)$ \Comment{Eq.~\eqref{eq:seg}}
    \EndIf
    \If{action token is \texttt{<ZOOM>} or \texttt{<SEGMENT>}}
        \State \textbf{Force} the model to emit an \texttt{Observation:} segment on $v_t$ until \texttt{<OBS\_END>}
        \State Append to context: $h_t \gets h_{t-1} \oplus a_t \oplus o_t$
    \Else
        \State \textbf{break} \Comment{reached \texttt{<ANSWER>}}
    \EndIf
\EndFor
\State \Return final answer span
\end{algorithmic}
\end{algorithm}

\begin{table*}[t]
\scriptsize
\centering
\setlength{\tabcolsep}{2pt}
\renewcommand{\arraystretch}{1.05}
\resizebox{\textwidth}{!}{
\begin{tabular}{@{}l c *{13}{c} @{}}
  \toprule
  \textbf{Model} & \textbf{Params}
  & \multicolumn{3}{c}{V$^{*}$}
  & \multicolumn{3}{c}{\textbf{HR Bench 4K}}
  & \multicolumn{3}{c}{\textbf{HR Bench 8K}}
  & \multicolumn{3}{c}{\textbf{MME-RW-Lite}}
  & \textbf{MMStar} \\
  &  & Attr & Spatial & Overall
     & FSP & FCP & Overall
     & FSP & FCP & Overall
     & Perc. & Reas. & All. &  All. \\
  \hline

  \multicolumn{15}{c}{\textbf{Proprietary models}} \\
  \hline
  GPT-4o & -
      & 71.30$^{*}$ & 69.74$^{*}$ & 70.68$^{*}$
      & 68.75$^{*}$ & 59.00$^{*}$ & 63.87$^{*}$
      & 60.75$^{*}$ & 60.00$^{*}$ & 60.38$^{*}$
      & 49.81 & 53.89 & 45.73 & 60.93 \\
  GPT-5 & -
      & 74.78 & 78.95 & 76.44
      & 78.00 & \underline{77.00} & 77.50
      & 70.50 & 72.75 & 71.62
      & 56.94 & \underline{55.47} & \underline{56.22} & \underline{76.47} \\
  GPT-5-nano & -
      & 58.26 & 72.37 & 63.87
      & 69.00 & 61.75 & 65.38
      & 65.75 & 61.50 & 63.63
      & 46.11 & 49.40 & 42.82 & 66.12 \\
  Gemini-2.5-pro & -
      & 84.35 & 71.05 & 79.06
      & 88.75 & \textbf{87.00} & \textbf{87.87}
      & 84.50 & \textbf{82.75} & \textbf{83.62}
      & 52.10 & 50.73 & 53.48 & \textbf{79.40} \\
  Gemini-2.5-flash & -
      & 84.35 & 73.68 & 80.10
      & 85.00 & \underline{77.00} & 81.00
      & 79.00 & \underline{73.50} & 76.25
      & 49.70 & 47.32 & 52.07 & 76.15 \\

  \hline
  \multicolumn{15}{c}{\textbf{Visual Reasoning Models}} \\
  \hline
  Pixel-Reasoner \cite{pixelreason} & 7B
      & - & - & 84.30$^{*}$
      & - & - & 74.00$^{*}$
      & - & - & 66.90$^{*}$
      & - & - & - & - \\
  Thyme \cite{zhang_thyme_2025} & 7B
      & - & - & 82.20$^{*}$
      & - & - & 77.00$^{*}$
      & - & - & 72.00$^{*}$
      & - & - & - & - \\
  LVR \cite{li2025latentvisualreasoning} & 7B
      & 81.7$^{*}$ & 79.0$^{*}$ & 80.6$^{*}$
      & - & - & -
      & - & - & -
      & - & - & - & - \\

  \hline
  \multicolumn{15}{c}{\textbf{Open-source models}} \\
  \hline
  LLaVA-OneVision-1.5 & 8B
      & 83.48 & 85.53 & 84.29
      & 72.00 & 46.50 & 59.25
      & 59.50 & 43.75 & 51.62
      & 38.69 & 36.53 & 37.61 & 60.73 \\
  Deepeyes & 7B
      & 84.35$^{*}$ & 81.58$^{*}$ & 83.03$^{*}$
      & 83.75$^{*}$ & 58.74$^{*}$ & 71.25$^{*}$
      & 77.00$^{*}$ & 53.25$^{*}$ & 65.13$^{*}$
      & 55.30 & 46.29 & 50.79 & 51.47 \\
  Monet-7B \cite{wang_monet_2025} & 7B
      & 83.48$^{*}$ & 82.89$^{*}$ & 83.25$^{*}$
      & 85.25$^{*}$ & 56.75$^{*}$ & 71.00$^{*}$
      & 79.75$^{*}$ & 56.25$^{*}$ & 68.00$^{*}$
      & 58.34$^{*}$ & 51.07$^{*}$ & 55.50$^{*}$ & - \\
  Qwen3-VL-Instruct & 235B$^\dagger$
      & 88.50 & 78.95 & 84.66
      & 90.50 & 71.75 & 81.13
      & 82.25 & 70.00 & 76.13
      & 56.30 & \textbf{60.78} & 51.82 & - \\
  Qwen3-VL-Instruct & 32B
      & 84.35 & 85.53 & 84.82
      & 90.50 & 69.00 & 79.75
      & 84.75 & 68.00 & \underline{76.38}
      & 45.07 & 51.07 & 39.07 & 70.93 \\
  Qwen3-VL-Thinking & 8B
      & 71.30 & 75.00 & 72.77
      & 81.00 & 60.00 & 70.50
      & 73.75 & 60.50 & 67.13
      & 44.05 & 34.80 & 39.43 & 70.80 \\
  Qwen3-VL-Instruct & 8B
      & 73.04 & 73.68 & 73.82
      & 90.50 & 57.75 & 74.13
      & 81.25 & 55.75 & 68.50
      & 49.02 & 34.27 & 41.64 & 65.07 \\

  \hline
  \multicolumn{15}{c}{\textbf{TikArt and ablations (ours)}} \\
  \hline
  TikArt & 8B
      & \textbf{91.30} & \textbf{86.84} & \textbf{89.53}
      & \underline{93.75} & 70.75 & \underline{82.25}
      & 84.50 & 68.25 & \underline{76.38}
      & \textbf{60.48} & 53.47 & \textbf{56.97} & 69.07 \\
  \rowcolor{gray!20}
  $\Delta$ (vs Qwen3-VL-8B-Instruct) &
      & \textcolor{ForestGreen}{+18.3} & \textcolor{ForestGreen}{+13.2} & \textcolor{ForestGreen}{+15.7}
      & \textcolor{ForestGreen}{+3.3}  & \textcolor{ForestGreen}{+13.0} & \textcolor{ForestGreen}{+8.1}
      & \textcolor{ForestGreen}{+3.3}  & \textcolor{ForestGreen}{+12.5} & \textcolor{ForestGreen}{+7.9}
      & \textcolor{ForestGreen}{+11.5} & \textcolor{ForestGreen}{+19.2} & \textcolor{ForestGreen}{+15.3} & \textcolor{ForestGreen}{+4.0} \\
  \rowcolor{gray!5}
  \quad w/o Observation &
      & \underline{89.57} & 81.58 & \underline{86.39}
      & \textbf{94.00} & 67.25 & 80.63
      & 85.25 & 64.75 & 75.00
      & 58.43 & 48.91 & 53.67 & 65.67 \\
  \rowcolor{gray!5}
  \quad w/o Zoom Action &
      & 76.32 & \underline{86.42} & 80.34
      & \underline{93.75} & 65.25 & 79.50
      & 85.25 & 64.25 & 74.75
      & 57.91 & 47.20 & 52.56 & 40.53 \\
  \rowcolor{gray!5}
  \quad w/o Segment Action &
      & 82.60 & 84.21 & 83.24
      & 93.25 & 59.75 & 76.50
      & \textbf{89.25} & 61.50 & 75.38
      & 53.51 & 47.87 & 50.69 & 66.93 \\
  \rowcolor{gray!5}
  \quad w/o GRPO Training &
      & 78.26 & 71.05 & 75.39
      & 92.00 & 70.25 & 81.13
      & 76.50 & 66.50 & 71.50
      & 50.65 & 46.55 & 48.60 & 66.38 \\
  \rowcolor{gray!5}
  \quad w/o RUR Reward &
      & 80.00 & 84.21 & 81.68
      & \textbf{94.00} & 65.00 & 79.50
      & \underline{87.25} & 62.50 & 74.88
      & \underline{60.05} & 50.67 & 55.36 & 67.53 \\

  \bottomrule
\end{tabular}
}
\caption{\label{tab:exp-vqa}
  Results on V$^{*}$, HR-Bench 4K/8K, MME-RealWorld-Lite, and MMStar. TikArt consistently improves over Qwen3-VL-8B-Instruct and narrows the gap to larger open-source and proprietary models. Values marked with $^*$ are from prior work; $^\dagger$ denotes MoE models with 235B total and 22B active parameters.
}

\end{table*}

\subsection{Reinforcement Learning with GRPO}

We optimize TikArt using a GRPO-style group-based policy optimization method. For each input $x$
(encoding the image, question, and formatting), we sample a group of $K$ rollouts
$\{\tau^{(k)}\}_{k=1}^K$ from the current policy.

\paragraph{Optimization target.}
The objective maximizes expected final reward:
\begin{equation}
\max_\theta\;\mathbb{E}_{\tau\sim\pi_\theta}\big[R_{\text{final}}(\tau)\big].
\label{eq:rl}
\end{equation}

\paragraph{Group-normalized advantage.}
For each input $x$, GRPO samples a group of $K$ rollouts $\{\tau^{(k)}\}_{k=1}^{K}$ and computes
their final rewards $\{R^{(k)}\}_{k=1}^{K}$, where $R^{(k)} \equiv R_{\text{final}}(\tau^{(k)})$.
We use a standardized group baseline to form advantages:
\begin{equation}
\mu_R \;=\; \frac{1}{K}\sum_{j=1}^{K} R^{(j)},
\qquad
\sigma_R \;=\; \sqrt{\frac{1}{K}\sum_{j=1}^{K}\left(R^{(j)}-\mu_R\right)^2},
\label{eq:group_stats}
\end{equation}
\begin{equation}
A^{(k)} \;=\; \frac{R^{(k)}-\mu_R}{\sigma_R+\epsilon},
\label{eq:group_adv_std}
\end{equation}
where $\epsilon>0$ is a small constant for numerical stability.

\paragraph{Why reward diversity matters in GRPO.}
GRPO standardizes final rewards within each prompt group (Eq.~\eqref{eq:group_adv_std}).
When the group rewards collapse to an identical value (e.g., all rollouts are incorrect and thus receive the same sparse task/action rewards),
the resulting advantages become zero and the policy update vanishes:
\begin{equation}
\forall k,\; R^{(k)} = c \;\Rightarrow\; \mu_R=c,\;\; A^{(k)}=0.
\label{eq:reward_collapse}
\end{equation}
To avoid this degeneracy---especially at early stages when most rollouts are wrong---we augment sparse outcome rewards with a dense, trajectory-sensitive signal (RUR; Sec.~\ref{sec:rur}) computed by a frozen evaluator.

\subsection{RUR: Dense Trajectory Reward via a Frozen LLM Evaluator}
\label{sec:rur}

Sparse outcome rewards are often insufficient for long-horizon tool-integrated rollouts: early in training, most candidates are incorrect, causing within-group reward ties under GRPO and weak policy updates.
We therefore introduce \textbf{Relative Uncertainty Reduction (RUR)}, a dense trajectory-validity reward computed by a frozen evaluator $M_{\mathrm{eval}}$.
In all experiments, we instantiate $M_{\mathrm{eval}}$ as Qwen3-VL-8B-Instruct and apply RUR to both discrete-answer reasoning tasks and segmentation-oriented settings through a unified textual evaluator interface.

\paragraph{Intuition.}
A TikArt trajectory is an evidence trail: each aperture and its mandatory Observation adds localized, grounded facts to the context.
RUR measures how much the trajectory prefix (strictly before \texttt{<ANSWER>}) increases the evaluator's confidence in the intended task target, encouraging evidence-building rollouts and helping break within-group reward ties.

\paragraph{Teacher-forced confidence of the task target.}
Given a context $c$ and a task target sequence
$y^{*}=(y^{*}_{1},\ldots,y^{*}_{|y^{*}|})$, we compute the average teacher-forced log-probability under $M_{\mathrm{eval}}$:
\begin{equation}
\mathcal{L}(c; y^{*})
=\frac{1}{|y^{*}|}\sum_{i=1}^{|y^{*}|}
\log p_{M_{\mathrm{eval}}}\!\left(y^{*}_{i}\mid c, y^{*}_{<i}\right).
\label{eq:rur_tf_logp}
\end{equation}
We map it back to probability scale via the per-token geometric mean:
\begin{equation}
p(c; y^{*})=\exp\!\big(\mathcal{L}(c; y^{*})\big).
\label{eq:rur_geom_prob}
\end{equation}
For discrete-answer tasks, $y^{*}$ is the ground-truth final answer sequence.
For segmentation and other non-text-output settings, $y^{*}$ is a short canonical text target derived from the supervision signal, which identifies the intended instance and coarse spatial attributes extracted from the ground-truth mask rather than encoding raw mask pixels.

\paragraph{Baseline vs. trajectory context.}
For a rollout $\tau$, let $\tau_{\mathrm{prefix}}$ denote the trajectory prefix strictly before \texttt{<ANSWER>} so as to avoid answer leakage.
We define
\begin{equation}
c_{0}=x,
\qquad
c_{\tau}=x\oplus\tau_{\mathrm{prefix}},
\label{eq:rur_contexts}
\end{equation}
and compute evaluator confidences
\begin{equation}
p_{\mathrm{base}}=p(c_{0}; y^{*}),
\qquad
p_{\mathrm{traj}}=p(c_{\tau}; y^{*}).
\label{eq:rur_base_traj_prob}
\end{equation}

\paragraph{Headroom-normalized confidence improvement.}
We define RUR as
\begin{equation}
\mathrm{RUR}(\tau)
=\frac{p_{\mathrm{traj}}-p_{\mathrm{base}}}{1-p_{\mathrm{base}}}.
\label{eq:rur_def}
\end{equation}
For numerical stability and boundedness, we use an $\epsilon$-stabilized denominator and symmetric clipping:
\begin{equation}
\mathrm{RUR}(\tau)=
\mathrm{clip}\!\left(
\frac{p_{\mathrm{traj}}-p_{\mathrm{base}}}{\max(\epsilon,\,1-p_{\mathrm{base}})},
-1,\,1\right),
\label{eq:rur_clip}
\end{equation}
where $\epsilon>0$ is a small constant.

\paragraph{Why the frozen evaluator helps.}
RUR is not an explicit KL penalty, but when the frozen evaluator is aligned with the base policy, it can play a KL-like anchoring role at the trajectory level.
To obtain a high RUR score, a rollout prefix must remain interpretable and useful under the evaluator, so repetitive strings, malformed traces, or semantically drifting tool use tend to receive weak or negative improvement.
We therefore use RUR as a practical stabilization signal together with task-native rewards, rather than as an evaluator-independent objective.

\subsection{Reward Design}
\label{sec:reward_design}

TikArt uses a composite reward that balances outcome correctness, purposeful aperture usage, and trajectory validity:
\begin{equation}
R_{\text{final}}(\tau)
= \beta_{1}\, R_{\text{task}}(\tau)
+ \beta_{2}\, R_{\text{action}}(\tau)
+ \beta_{3}\, \mathrm{RUR}(\tau),
\label{eq:reward}
\end{equation}
where $\beta_1,\beta_2,\beta_3>0$ control the trade-offs.
Unless otherwise stated, we use \emph{task : action : RUR} $= 0.8 : 1.2 : 1.0$.

\paragraph{Task reward $R_{\text{task}}$.}
We compute $R_{\text{task}}$ according to the supervision type.
For \textbf{VQA, chart understanding, and other discrete QA tasks}, the task reward is a binary correctness signal
\begin{equation}
R_{\text{acc}}\in\{0,1\},
\end{equation}
computed from whether the extracted final answer matches the reference answer under task-compatible normalization or fixed judging rules.
For \textbf{visual math}, we parse the final boxed answer and apply rule-based equivalence checking to obtain a binary correctness signal.
For \textbf{segmentation}, we use a continuous reward and set
\begin{equation}
R_{\text{seg}}=(1-\alpha)\,R_{\text{IoU}}+\alpha\,R_{\text{S}},
\label{eq:seg_reward}
\end{equation}
where $R_{\text{IoU}}$ measures mask overlap quality and $R_{\text{S}}$ denotes the auxiliary segmentation score.
Accordingly,
\begin{equation}
R_{\text{task}}(\tau)=
\begin{cases}
R_{\text{acc}}(\tau), & \text{for discrete-answer tasks},\\
R_{\text{seg}}(\tau), & \text{for segmentation tasks}.
\end{cases}
\end{equation}
Importantly, $R_{\text{task}}$ specifies only the \emph{task-dependent outcome term}.
Both discrete-answer and segmentation trajectories additionally receive the shared trajectory-validity term $\mathrm{RUR}(\tau)$ through Eq.~\eqref{eq:reward}.

\paragraph{Format penalty.}
To discourage malformed traces and reward exploitation through invalid formatting, we add a small negative shaping term
\begin{equation}
R_{\text{fmt}}(\tau)\in\{-1,0\},
\end{equation}
where $R_{\text{fmt}}=-1$ if structural tags are unbalanced or the final answer exceeds a hard length cap; otherwise $R_{\text{fmt}}=0$.
For discrete-answer tasks, we fold this term into the task reward:
\begin{equation}
R_{\text{task}}(\tau)\leftarrow R_{\text{acc}}(\tau)+\lambda_{\text{fmt}}\,R_{\text{fmt}}(\tau),
\label{eq:task_with_fmt}
\end{equation}
with a small $\lambda_{\text{fmt}}>0$.

\paragraph{Action reward $R_{\text{action}}$.}
$R_{\text{action}}(\tau)\in\{0,1\}$ encourages \emph{purposeful} aperture usage.
Let $n_{\mathrm{ap}}(\tau)$ denote the number of successfully executed aperture actions within trajectory $\tau$.
We assign action reward only when the policy both performs at least one successful aperture and reaches a task-compatible successful output:
\begin{equation}
R_{\text{action}}(\tau)
=\mathbb{I}\!\left[n_{\mathrm{ap}}(\tau)\ge 1\right]\cdot \mathbb{I}\!\left[S(\tau)=1\right],
\label{eq:action_reward}
\end{equation}
where $S(\tau)$ denotes task success under the corresponding task-specific evaluation rule.
This conservative ``look-and-win'' design prevents the model from invoking apertures merely to lengthen trajectories.
Correct zero-aperture trajectories can still succeed through $R_{\text{task}}$ alone, while $R_{\text{action}}$ only provides additional credit for successful evidence acquisition.
Thus, the binary action term discourages gratuitous tool use, and RUR restores denser credit assignment for informative but unsuccessful explorations.

\begin{figure*}[!tbp]
    \centering
    \begin{subfigure}[b]{0.32\linewidth}
        \centering
        \includegraphics[width=\linewidth]{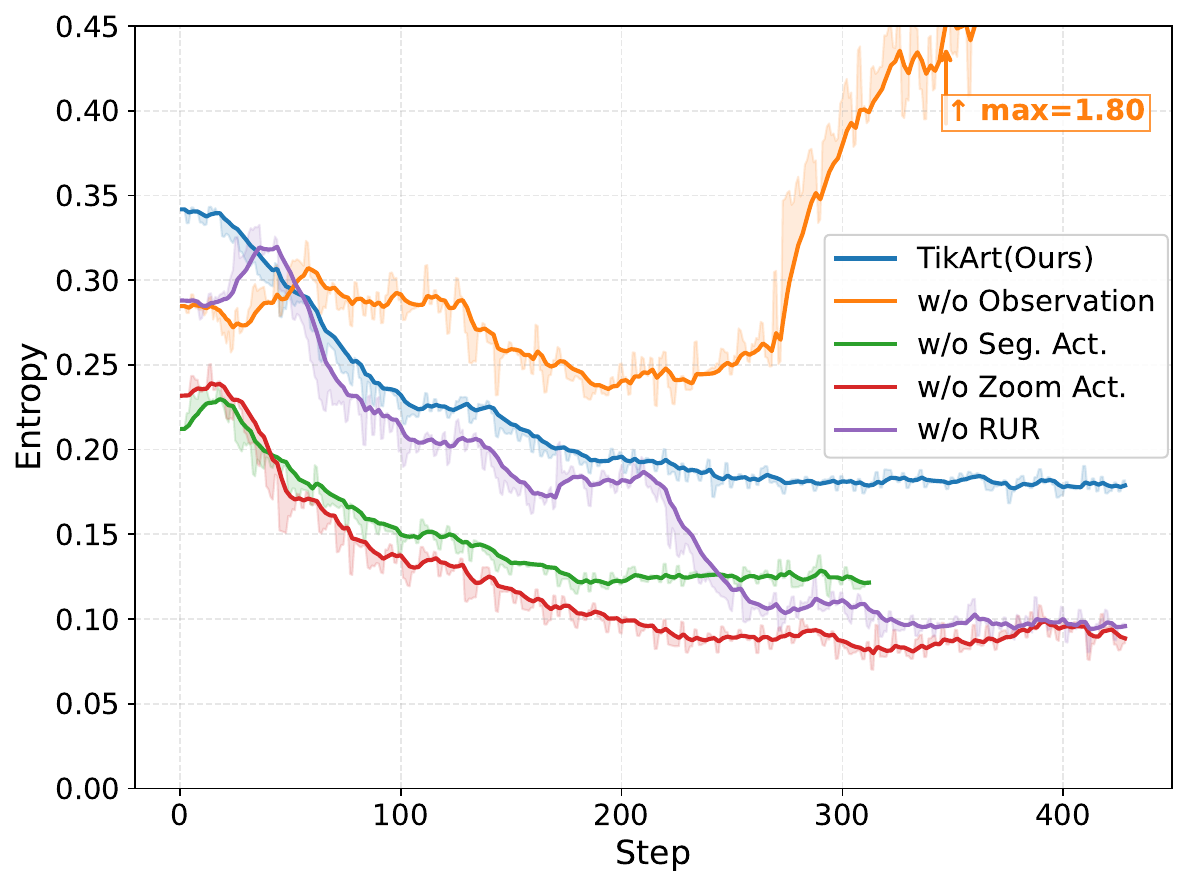}
        \caption{Policy entropy}
        \label{fig:entropy_curves}
    \end{subfigure}
    \hfill
    \begin{subfigure}[b]{0.32\linewidth}
        \centering
        \includegraphics[width=\linewidth]{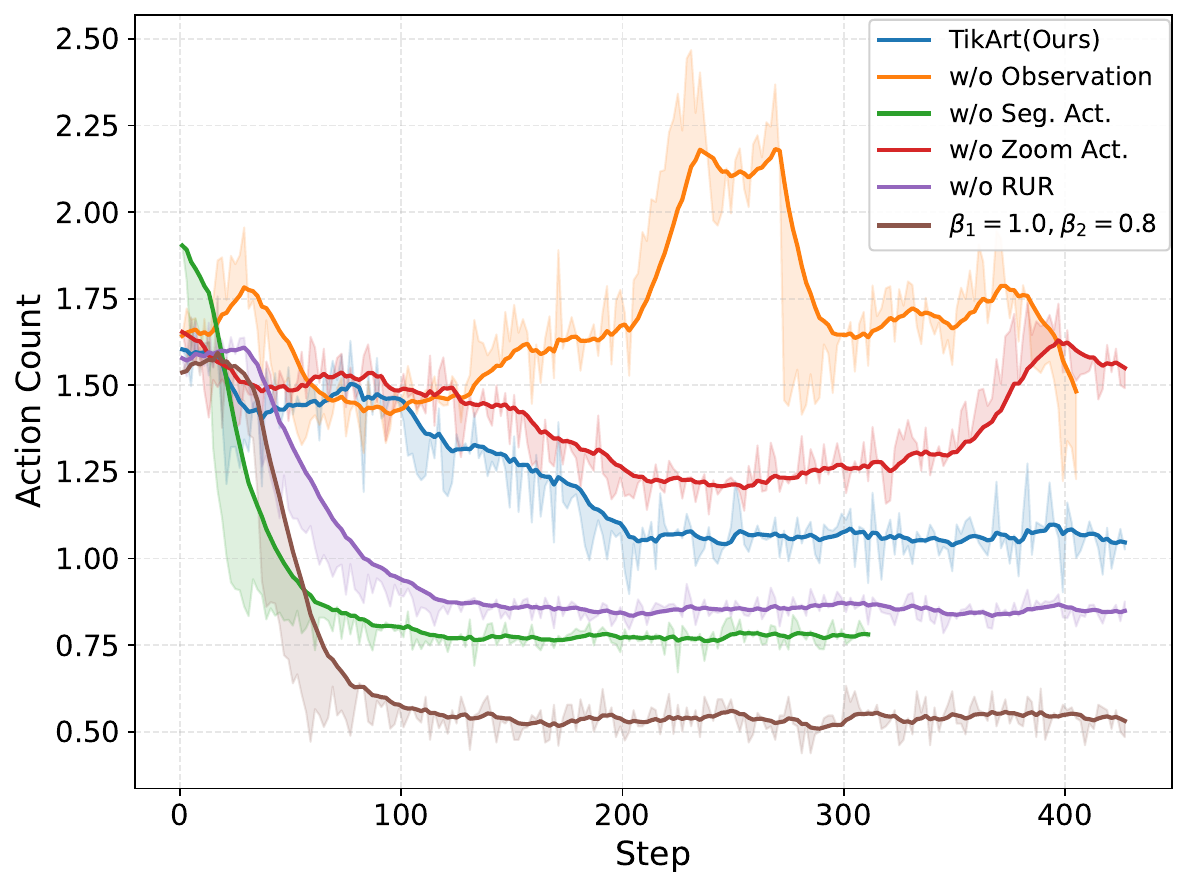}
        \caption{Aperture action count}
        \label{fig:action_count_curves}
    \end{subfigure}
    \hfill
    \begin{subfigure}[b]{0.32\linewidth}
        \centering
        \includegraphics[width=\linewidth]{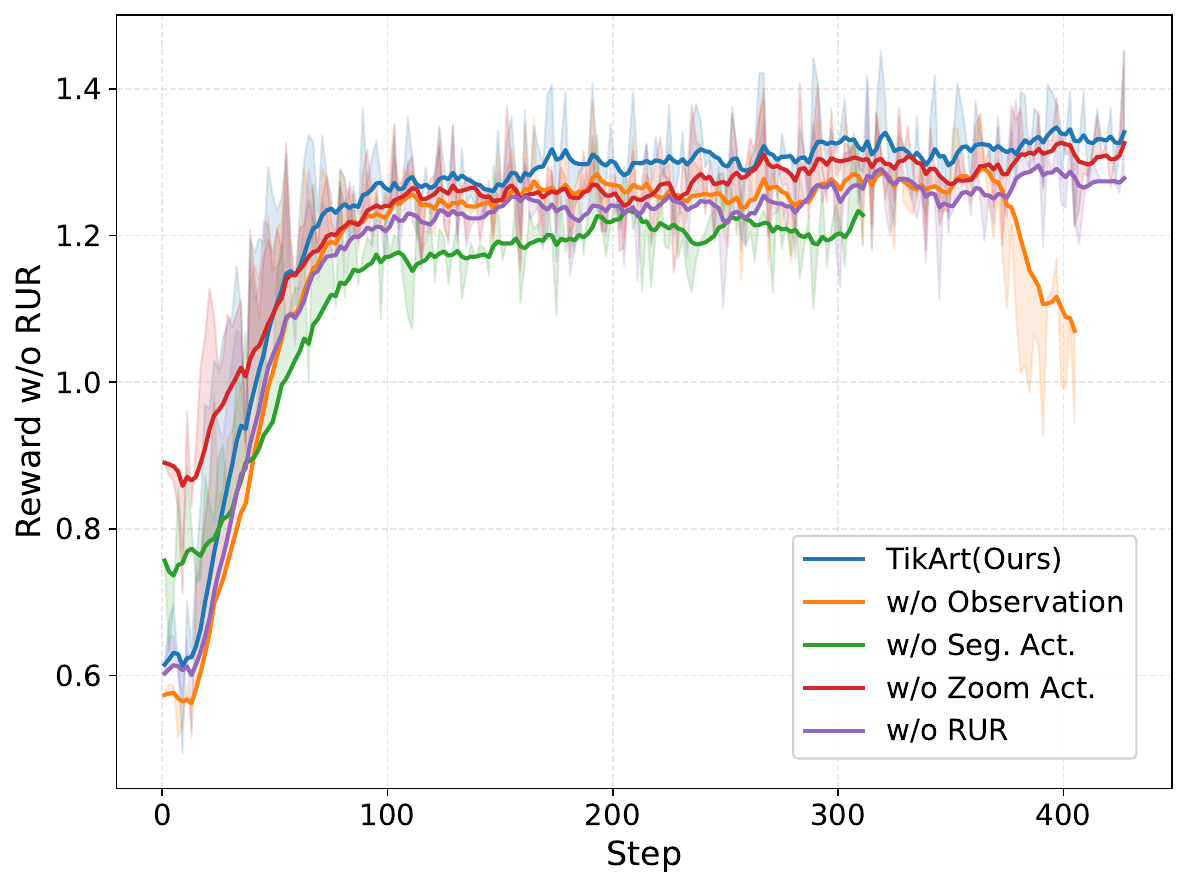}
        \caption{Group reward (w/o RUR)}
        \label{fig:reward_curves}
    \end{subfigure}

    \caption{
        Training dynamics of TikArt and GRPO-based variants.
        We plot (a) policy entropy, (b) the average number of aperture actions per trajectory, and (c) the group reward excluding RUR, i.e., aggregated from task and action terms only.
        TikArt converges to a stable policy with moderate entropy and reasonable aperture usage, while removing Observation leads to higher entropy, uncontrolled aperture actions, and degraded reward.
    }
    \label{fig:training_dynamics}
\end{figure*}

\section{Experiments and Ablations}

\subsection{Experimental Setup}
\label{sec:exp_setup}

\paragraph{Benchmarks.}
We evaluate TikArt on four task families: (1) high-resolution fine-grained reasoning (V$^{*}$, HR-Bench 4K/8K, and MME-RealWorld-Lite), (2) general multimodal understanding (MMStar), (3) referring segmentation (RefCOCO), and (4) reasoning-oriented segmentation (ReasonSeg).

\paragraph{Models and evaluation.}
TikArt is built on Qwen3-VL-8B-Instruct and compared with its backbone, stronger open-source models (e.g., LLaVA-OneVision-1.5-8B, Qwen3-VL-32B/235B), and proprietary systems such as GPT-4o/GPT-5 and Gemini-2.5.
All models are evaluated under matched task prompts; for tool-enabled variants, we use the same tool-control prompt and the same preprocessed images.
RUR uses a single frozen evaluator, Qwen3-VL-8B-Instruct, across all tasks.
Final answers are scored by rule-based normalization or fixed task-specific judging rules, with prompts and extraction details deferred to the supplementary material.

\paragraph{Training settings.}
We train on one node with 8$\times$NVIDIA A100 (40GB) GPUs for one epoch.
GRPO uses $K{=}8$ rollouts per prompt, $\beta{=}0$ (no KL penalty), and an upper clipping threshold $\epsilon_{\text{high}}{=}0.28$ on the policy-ratio term, with maximum completion length 4096, learning rate $1\mathrm{e}{-6}$, and warmup ratio 0.05.
Segmentation apertures call a frozen SAM2 (\texttt{sam2.1\_hiera\_large}) module.
Unless otherwise stated, all ablations use the same backbone, evaluator, segmentation backend, and rollout configuration.

\paragraph{Training data and curriculum.}
Training follows a two-stage curriculum over 57K instances in total.
The corpus comprises 18.8K examples from the V$^{*}$~\cite{wu2024v} training set (33.0\%), 12.0K chart-centric examples from ArxivQA~\cite{arxivqa} (21.1\%), 9.2K reasoning examples from ThinkLite-VL~\cite{thinklitevl} (16.1\%), and 17.0K segmentation-oriented examples (29.8\%) drawn from COCO2017, ReasonSeg (train), and ADE20K.
We first use 2,000 COCO2017 samples as a segmentation warm-up stage, and then perform multi-task GRPO on the remaining 55K instances.
We verify that no evaluation example from V$^{*}$, HR-Bench, MME-RealWorld-Lite, RefCOCO, or ReasonSeg test splits is included in training.

\paragraph{Fairness and implementation notes.}
All TikArt variants share the same TAO formatting, image preprocessing pipeline, and external segmentation backend unless that component is explicitly ablated.
For proprietary APIs, we use the same preprocessed images and prompt format, but these systems do not execute TikArt's interactive aperture stack; such comparisons should therefore be viewed as broad capability references rather than strict tool-parity controls.
We will release the aperture controller, prompts, and task-scoring scripts.

\subsection{High-Resolution and Real-World Benchmarks}

Table~\ref{tab:exp-vqa} reports results on V$^{*}$, HR-Bench 4K/8K, MME-RealWorld-Lite, and MMStar.
On V$^{*}$, TikArt-8B improves both attribute-centric and spatial reasoning, yielding +18.3 on \textit{Attr.} and +13.2 on \textit{Spatial}, for an overall gain of +15.7 over Qwen3-VL-8B-Instruct.
On HR-Bench, gains are particularly pronounced on fine-compositional perception (\textit{FCP}), e.g., +13.0 (4K) and +12.5 (8K), suggesting that aperture-guided observation better recovers small or cluttered evidence that is under-represented in a single global encoding.
On MME-RealWorld-Lite, TikArt substantially boosts \textit{Reasoning} (+19.2) alongside \textit{Perception} (+11.5), indicating that the learned RoI policy transfers beyond synthetic high-resolution settings to real-world scenarios where multi-step evidence accumulation is required.
Overall, TikArt narrows the gap to much larger open-source and proprietary systems while remaining at the 8B scale.

\begin{table}[t]

\small

\centering

\setlength{\tabcolsep}{3pt}

\begin{tabular}{@{}lcccccc@{}}

  \toprule

  \textbf{Model} & \textbf{Params}
  & \multicolumn{3}{c}{\textbf{RefCOCO}} 
  & \multicolumn{2}{c} {\textbf{ReasonSeg-test}} \\
  &  & val & testa & testb & giou & ciou \\

\hline
  Grounded SAM & -   & - & - & - & 21.3$^{*}$ & 16.4$^{*}$ \\
  
  LISA & 13B   & 74.9$^{*}$ & 79.1$^{*}$ & 72.4$^{*}$ & 61.3$^{*}$ & 62.2$^{*}$ \\
  SegR1 & 7B & 74.3$^{*}$ & 78.7$^{*}$ & 67.6$^{*}$ & 56.7$^{*}$ & 53.7$^{*}$ \\
  VisionReasoner & 7B & - & - & - & 63.6$^{*}$ & - \\
  SAM-R1 & 7B & \textbf{79.2}$^{*}$ & - & - & 60.2$^{*}$ & 54.3$^{*}$ \\

  \rowcolor{gray!15}
  TikArt & 8B   & 77.1 & 79.6 & 69.1 & \textbf{73.8} & \textbf{73.2}\\
  \rowcolor{gray!15}
  \quad w/o RUR Reward  &  & 77.3 &  78.6 & 67.9 & 71.2 & 69.5\\
  \rowcolor{gray!15}
  \quad w/o Observation  &  & 74.1 &  75.4 & 66.1 & 70.0 & 68.7\\
  \rowcolor{gray!15}
  \quad w/o Zoom Action &     & 74.9 & \textbf{82.3} & \textbf{73.3} & 67.2 & 65.8 \\
  \rowcolor{gray!15}
  \quad w/o Aperture    &     & 73.8 & 74.2 & 64.2 & 69.6 & 68.1\\
  \bottomrule
\end{tabular}

\caption{\label{tab:exp-segments}
  Results on referring segmentation and reasoning-oriented segmentation benchmarks. TikArt brings ReasonSeg performance up to or beyond RL segmentation baselines while retaining strong RefCOCO scores. We also include \textit{w/o RUR Reward} to quantify the contribution of the shared trajectory-validity reward in segmentation settings. Values marked with $\ast$ are taken from prior work~\cite{lai2024lisa,Seg-R1,huang2025samr1leveragingsamreward}. The best result in each column is shown in bold. TikArt \textit{w/o Aperture} disables aperture control; \textit{w/o Zoom Action} disallows the zoom action; \textit{w/o Observation} removes the observation module.
}

\end{table}

\subsection{Segmentation}

Table~\ref{tab:exp-segments} compares TikArt on RefCOCO and ReasonSeg.
TikArt-8B achieves strong ReasonSeg performance (73.8 gIoU / 73.2 cIoU), outperforming prior RL-based segmentation baselines such as SegR1 and SAM-R1 by a large margin, while maintaining competitive RefCOCO accuracy (77.1/79.6/69.1 on val/testA/testB).
The \textit{w/o RUR Reward} ablation also degrades segmentation performance (71.2 gIoU / 69.5 cIoU on ReasonSeg), showing that RUR benefits not only discrete-answer reasoning but also mask-oriented grounding by rewarding evidence-building trajectory prefixes.
These results suggest that the same aperture policy learned for fine-grained VQA transfers naturally to pixel-level grounding.
Because all TikArt variants in Table~\ref{tab:exp-segments} use the same frozen SAM2 backend, the gains within these ablations primarily reflect policy and trajectory-learning effects rather than changes in the segmentation engine itself.

\subsection{Ablation Studies}
\label{sec:ablation}

We ablate the main components in Table~\ref{tab:exp-vqa} and Table~\ref{tab:exp-segments}.
Removing RUR (\textit{w/o RUR Reward}) consistently hurts both discrete-answer reasoning and segmentation, indicating that trajectory-level validity complements sparse task and action rewards.
Removing \textbf{Observation} causes the clearest instability: Fig.~\ref{fig:training_dynamics} shows higher policy entropy, heavier-tailed aperture usage, and weaker reward, confirming that explicit observations are a learning interface rather than merely explanatory text.
Action ablations reveal functional specialization rather than a universal dominance relation: \textit{w/o Segment Action} is weaker on thin, irregular, or cluttered targets, whereas \textit{w/o Zoom Action} is less effective on structured evidence such as charts, panels, and tables.
The prompt-only baseline (\textit{w/o GRPO}) still improves several benchmarks, showing that the TAO interface itself is useful, but full RL remains important for learning when to act, where to look, and how to turn local evidence into stable observations.

\paragraph{Ablation protocol.}
Unless otherwise stated, all variants share the optimization settings in Sec.~\ref{sec:exp_setup}.
Component ablations that preserve the full action space (e.g., \textit{w/o RUR Reward} and \textit{w/o Observation}) are trained on the full multi-task mixture.
Action ablations use the largest compatible subset of the corpus, so they should be interpreted primarily as evidence of functional specialization rather than strictly matched replacements.

\section{Conclusion}
\label{sec:conclusion}

We presented \textbf{TikArt}, an aperture-guided agent for fine-grained visual reasoning that integrates iterative perception into the reasoning process.
TikArt formulates multimodal reasoning as sequential evidence acquisition over regions of interest and follows a Think--Aperture--Observe loop with two complementary actions: \textbf{Zoom} for structured local inspection and \textbf{Segment} for object-centric perception on irregular targets.

A central design choice is the \textbf{mandatory Observation} step after every aperture action.
By writing local evidence back into text, TikArt converts transient visual inspection into persistent and auditable memory, strengthening the link between perception, intermediate reasoning, and final prediction.
We further introduce \textbf{RUR}, a dense reward computed by a frozen evaluator that stabilizes evidence-building trajectories under GRPO-style reinforcement learning across both reasoning and segmentation settings.

Across high-resolution reasoning, real-world multimodal understanding, and both referring and reasoning-oriented segmentation benchmarks, TikArt consistently improves over its backbone and yields interpretable aperture trajectories.
These results suggest that aperture-guided observation is a practical and effective interface for fine-grained multimodal reasoning and pixel-level grounding.

\bibliography{segment,Think_with_image,dataset_benchmark,custom}


\begin{thebibliography}{32}


\ifx \showCODEN    \undefined \def \showCODEN     #1{\unskip}     \fi
\ifx \showISBNx    \undefined \def \showISBNx     #1{\unskip}     \fi
\ifx \showISBNxiii \undefined \def \showISBNxiii  #1{\unskip}     \fi
\ifx \showISSN     \undefined \def \showISSN      #1{\unskip}     \fi
\ifx \showLCCN     \undefined \def \showLCCN      #1{\unskip}     \fi
\ifx \shownote     \undefined \def \shownote      #1{#1}          \fi
\ifx \showarticletitle \undefined \def \showarticletitle #1{#1}   \fi
\ifx \showURL      \undefined \def \showURL       {\relax}        \fi
\providecommand\bibfield[2]{#2}
\providecommand\bibinfo[2]{#2}
\providecommand\natexlab[1]{#1}
\providecommand\showeprint[2][]{arXiv:#2}

\bibitem[Bai et~al\mbox{.}(2023)]%
        {bai2023qwen}
\bibfield{author}{\bibinfo{person}{Jinze Bai}, \bibinfo{person}{Shuai Bai},
  \bibinfo{person}{Yunfei Chu}, \bibinfo{person}{Zeyu Cui},
  \bibinfo{person}{Kai Dang}, \bibinfo{person}{Xiaodong Deng},
  \bibinfo{person}{Yang Fan}, \bibinfo{person}{Wenbin Ge}, \bibinfo{person}{Yu
  Han}, \bibinfo{person}{Fei Huang}, {et~al\mbox{.}}}
  \bibinfo{year}{2023}\natexlab{}.
\newblock \showarticletitle{Qwen technical report}.
\newblock \bibinfo{journal}{\emph{arXiv preprint arXiv:2309.16609}}
  (\bibinfo{year}{2023}).
\newblock


\bibitem[Bai et~al\mbox{.}(2025)]%
        {bai2025qwen3vltechnicalreport}
\bibfield{author}{\bibinfo{person}{Shuai Bai}, \bibinfo{person}{Yuxuan Cai},
  \bibinfo{person}{Ruizhe Chen}, \bibinfo{person}{Keqin Chen},
  \bibinfo{person}{Xionghui Chen}, \bibinfo{person}{Zesen Cheng},
  \bibinfo{person}{Lianghao Deng}, \bibinfo{person}{Wei Ding},
  \bibinfo{person}{Chang Gao}, \bibinfo{person}{Chunjiang Ge},
  \bibinfo{person}{Wenbin Ge}, \bibinfo{person}{Zhifang Guo},
  \bibinfo{person}{Qidong Huang}, \bibinfo{person}{Jie Huang},
  \bibinfo{person}{Fei Huang}, \bibinfo{person}{Binyuan Hui},
  \bibinfo{person}{Shutong Jiang}, \bibinfo{person}{Zhaohai Li},
  \bibinfo{person}{Mingsheng Li}, \bibinfo{person}{Mei Li},
  \bibinfo{person}{Kaixin Li}, \bibinfo{person}{Zicheng Lin},
  \bibinfo{person}{Junyang Lin}, \bibinfo{person}{Xuejing Liu},
  \bibinfo{person}{Jiawei Liu}, \bibinfo{person}{Chenglong Liu},
  \bibinfo{person}{Yang Liu}, \bibinfo{person}{Dayiheng Liu},
  \bibinfo{person}{Shixuan Liu}, \bibinfo{person}{Dunjie Lu},
  \bibinfo{person}{Ruilin Luo}, \bibinfo{person}{Chenxu Lv},
  \bibinfo{person}{Rui Men}, \bibinfo{person}{Lingchen Meng},
  \bibinfo{person}{Xuancheng Ren}, \bibinfo{person}{Xingzhang Ren},
  \bibinfo{person}{Sibo Song}, \bibinfo{person}{Yuchong Sun},
  \bibinfo{person}{Jun Tang}, \bibinfo{person}{Jianhong Tu},
  \bibinfo{person}{Jianqiang Wan}, \bibinfo{person}{Peng Wang},
  \bibinfo{person}{Pengfei Wang}, \bibinfo{person}{Qiuyue Wang},
  \bibinfo{person}{Yuxuan Wang}, \bibinfo{person}{Tianbao Xie},
  \bibinfo{person}{Yiheng Xu}, \bibinfo{person}{Haiyang Xu},
  \bibinfo{person}{Jin Xu}, \bibinfo{person}{Zhibo Yang},
  \bibinfo{person}{Mingkun Yang}, \bibinfo{person}{Jianxin Yang},
  \bibinfo{person}{An Yang}, \bibinfo{person}{Bowen Yu}, \bibinfo{person}{Fei
  Zhang}, \bibinfo{person}{Hang Zhang}, \bibinfo{person}{Xi Zhang},
  \bibinfo{person}{Bo Zheng}, \bibinfo{person}{Humen Zhong},
  \bibinfo{person}{Jingren Zhou}, \bibinfo{person}{Fan Zhou},
  \bibinfo{person}{Jing Zhou}, \bibinfo{person}{Yuanzhi Zhu}, {and}
  \bibinfo{person}{Ke Zhu}.} \bibinfo{year}{2025}\natexlab{}.
\newblock \bibinfo{title}{Qwen3-VL Technical Report}.
\newblock
\showeprint[arxiv]{2511.21631}~[cs.CV]
\urldef\tempurl%
\url{https://arxiv.org/abs/2511.21631}
\showURL{%
\tempurl}


\bibitem[Choudhury(2025)]%
        {choudhury2025agentprm}
\bibfield{author}{\bibinfo{person}{Sanjiban Choudhury}.}
  \bibinfo{year}{2025}\natexlab{}.
\newblock \showarticletitle{Process Reward Models for LLM Agents: Practical
  Framework and Directions}.
\newblock \bibinfo{journal}{\emph{arXiv preprint arXiv:2502.10325}}
  (\bibinfo{year}{2025}).
\newblock


\bibitem[Comanici et~al\mbox{.}(2025)]%
        {gemini-2.5}
\bibfield{author}{\bibinfo{person}{Gheorghe Comanici}, \bibinfo{person}{Eric
  Bieber}, \bibinfo{person}{Mike Schaekermann}, \bibinfo{person}{Ice Pasupat},
  \bibinfo{person}{Noveen Sachdeva}, \bibinfo{person}{Inderjit Dhillon},
  \bibinfo{person}{Marcel Blistein}, \bibinfo{person}{Ori Ram},
  \bibinfo{person}{Dan Zhang}, \bibinfo{person}{Evan Rosen}, {et~al\mbox{.}}}
  \bibinfo{year}{2025}\natexlab{}.
\newblock \showarticletitle{Gemini 2.5: Pushing the frontier with advanced
  reasoning, multimodality, long context, and next generation agentic
  capabilities}.
\newblock \bibinfo{journal}{\emph{arXiv preprint arXiv:2507.06261}}
  (\bibinfo{year}{2025}).
\newblock


\bibitem[Fan et~al\mbox{.}(2026)]%
        {fan2026gfmllm}
\bibfield{author}{\bibinfo{person}{Zhendong Fan}, \bibinfo{person}{Cheng
  Zhang}, \bibinfo{person}{Jinyang Gao}, \bibinfo{person}{Hao Wei},
  \bibinfo{person}{Hongbo Gao}, \bibinfo{person}{Tao Xie},
  \bibinfo{person}{Ruifeng Li}, {and} \bibinfo{person}{Lijun Zhao}.}
  \bibinfo{year}{2026}\natexlab{}.
\newblock \showarticletitle{GFMLLM: Enhance Multi-Modal Large Language Model
  for Global and Fine-Grained Visual Spatial Perception}.
\newblock \bibinfo{journal}{\emph{Expert Systems with Applications}}
  \bibinfo{volume}{299} (\bibinfo{year}{2026}), \bibinfo{pages}{130239}.
\newblock
\href{https://doi.org/10.1016/j.eswa.2025.130239}{doi:\nolinkurl{10.1016/j.eswa.2025.130239}}


\bibitem[Feng et~al\mbox{.}(2026)]%
        {feng2026rewardmap}
\bibfield{author}{\bibinfo{person}{Sicheng Feng}, \bibinfo{person}{Kaiwen Tuo},
  \bibinfo{person}{Song Wang}, \bibinfo{person}{Lingdong Kong},
  \bibinfo{person}{Jianke Zhu}, {and} \bibinfo{person}{Huan Wang}.}
  \bibinfo{year}{2026}\natexlab{}.
\newblock \bibinfo{title}{RewardMap: Tackling Sparse Rewards in Fine-grained
  Visual Reasoning via Multi-Stage Reinforcement Learning}.
\newblock
\showeprint[arxiv]{2510.02240}~[cs.CV]
\urldef\tempurl%
\url{https://arxiv.org/abs/2510.02240}
\showURL{%
\tempurl}


\bibitem[He et~al\mbox{.}(2026)]%
        {he2026finer1}
\bibfield{author}{\bibinfo{person}{Hulingxiao He}, \bibinfo{person}{Zijun
  Geng}, {and} \bibinfo{person}{Yuxin Peng}.} \bibinfo{year}{2026}\natexlab{}.
\newblock \showarticletitle{Fine-R1: Make Multi-modal LLMs Excel in
  Fine-Grained Visual Recognition by Chain-of-Thought Reasoning}. In
  \bibinfo{booktitle}{\emph{International Conference on Learning
  Representations (ICLR)}}.
\newblock
\urldef\tempurl%
\url{https://openreview.net/forum?id=kyzHM557gE}
\showURL{%
\tempurl}


\bibitem[Hong et~al\mbox{.}(2025)]%
        {hong_deepeyesv2_2025}
\bibfield{author}{\bibinfo{person}{Jack Hong}, \bibinfo{person}{Chenxiao Zhao},
  \bibinfo{person}{ChengLin Zhu}, \bibinfo{person}{Weiheng Lu},
  \bibinfo{person}{Guohai Xu}, {and} \bibinfo{person}{Xing Yu}.}
  \bibinfo{year}{2025}\natexlab{}.
\newblock \showarticletitle{{DeepEyesV2}: {Toward} {Agentic} {Multimodal}
  {Model}}.
\newblock  (\bibinfo{date}{Nov.} \bibinfo{year}{2025}).
\newblock
\href{https://doi.org/10.48550/arXiv.2511.05271}{doi:\nolinkurl{10.48550/arXiv.2511.05271}}
\newblock
\shownote{arXiv:2511.05271 [cs]}.


\bibitem[Hu et~al\mbox{.}(2024)]%
        {skecth}
\bibfield{author}{\bibinfo{person}{Yushi Hu}, \bibinfo{person}{Weijia Shi},
  \bibinfo{person}{Xingyu Fu}, \bibinfo{person}{Dan Roth},
  \bibinfo{person}{Mari Ostendorf}, \bibinfo{person}{Luke Zettlemoyer},
  \bibinfo{person}{Noah~A Smith}, {and} \bibinfo{person}{Ranjay Krishna}.}
  \bibinfo{year}{2024}\natexlab{}.
\newblock \bibinfo{title}{Visual Sketchpad: Sketching as a Visual Chain of
  Thought for Multimodal Language Models}.
\newblock
\showeprint[arxiv]{2406.09403}~[cs.CV]
\urldef\tempurl%
\url{https://arxiv.org/abs/2406.09403}
\showURL{%
\tempurl}


\bibitem[Huang et~al\mbox{.}(2025)]%
        {huang2025samr1leveragingsamreward}
\bibfield{author}{\bibinfo{person}{Jiaqi Huang}, \bibinfo{person}{Zunnan Xu},
  \bibinfo{person}{Jun Zhou}, \bibinfo{person}{Ting Liu},
  \bibinfo{person}{Yicheng Xiao}, \bibinfo{person}{Mingwen Ou},
  \bibinfo{person}{Bowen Ji}, \bibinfo{person}{Xiu Li}, {and}
  \bibinfo{person}{Kehong Yuan}.} \bibinfo{year}{2025}\natexlab{}.
\newblock \bibinfo{title}{SAM-R1: Leveraging SAM for Reward Feedback in
  Multimodal Segmentation via Reinforcement Learning}.
\newblock
\showeprint[arxiv]{2505.22596}~[cs.CV]
\urldef\tempurl%
\url{https://arxiv.org/abs/2505.22596}
\showURL{%
\tempurl}


\bibitem[Kirillov et~al\mbox{.}(2023)]%
        {kirillov2023segment}
\bibfield{author}{\bibinfo{person}{Alexander Kirillov}, \bibinfo{person}{Eric
  Mintun}, \bibinfo{person}{Nikhila Ravi}, \bibinfo{person}{Hanzi Mao},
  \bibinfo{person}{Chloe Rolland}, \bibinfo{person}{Laura Gustafson},
  \bibinfo{person}{Tete Xiao}, \bibinfo{person}{Spencer Whitehead},
  \bibinfo{person}{Alexander~C. Berg}, \bibinfo{person}{Wan-Yen Lo},
  \bibinfo{person}{Piotr Dollár}, {and} \bibinfo{person}{Ross Girshick}.}
  \bibinfo{year}{2023}\natexlab{}.
\newblock \bibinfo{title}{Segment Anything}.
\newblock
\showeprint[arxiv]{2304.02643}~[cs.CV]
\urldef\tempurl%
\url{https://arxiv.org/abs/2304.02643}
\showURL{%
\tempurl}


\bibitem[Lai et~al\mbox{.}(2024)]%
        {lai2024lisa}
\bibfield{author}{\bibinfo{person}{Xin Lai}, \bibinfo{person}{Zhuotao Tian},
  \bibinfo{person}{Yukang Chen}, \bibinfo{person}{Yanwei Li},
  \bibinfo{person}{Yuhui Yuan}, \bibinfo{person}{Shu Liu}, {and}
  \bibinfo{person}{Jiaya Jia}.} \bibinfo{year}{2024}\natexlab{}.
\newblock \showarticletitle{Lisa: Reasoning segmentation via large language
  model}. In \bibinfo{booktitle}{\emph{Proceedings of the IEEE/CVF Conference
  on Computer Vision and Pattern Recognition}}. \bibinfo{pages}{9579--9589}.
\newblock


\bibitem[Li et~al\mbox{.}(2025)]%
        {li2025latentvisualreasoning}
\bibfield{author}{\bibinfo{person}{Bangzheng Li}, \bibinfo{person}{Ximeng Sun},
  \bibinfo{person}{Jiang Liu}, \bibinfo{person}{Ze Wang},
  \bibinfo{person}{Jialian Wu}, \bibinfo{person}{Xiaodong Yu},
  \bibinfo{person}{Hao Chen}, \bibinfo{person}{Emad Barsoum},
  \bibinfo{person}{Muhao Chen}, {and} \bibinfo{person}{Zicheng Liu}.}
  \bibinfo{year}{2025}\natexlab{}.
\newblock \bibinfo{title}{Latent Visual Reasoning}.
\newblock
\showeprint[arxiv]{2509.24251}~[cs.CV]
\urldef\tempurl%
\url{https://arxiv.org/abs/2509.24251}
\showURL{%
\tempurl}


\bibitem[Li et~al\mbox{.}(2024)]%
        {arxivqa}
\bibfield{author}{\bibinfo{person}{Lei Li}, \bibinfo{person}{Yuqi Wang},
  \bibinfo{person}{Runxin Xu}, \bibinfo{person}{Peiyi Wang},
  \bibinfo{person}{Xiachong Feng}, \bibinfo{person}{Lingpeng Kong}, {and}
  \bibinfo{person}{Qi Liu}.} \bibinfo{year}{2024}\natexlab{}.
\newblock \showarticletitle{Multimodal arxiv: A dataset for improving
  scientific comprehension of large vision-language models}. In
  \bibinfo{booktitle}{\emph{Proceedings of the 62nd Annual Meeting of the
  Association for Computational Linguistics (Volume 1: Long Papers)}}.
  \bibinfo{pages}{14369--14387}.
\newblock


\bibitem[Liu et~al\mbox{.}(2025)]%
        {liu2025istar}
\bibfield{author}{\bibinfo{person}{Xiaoqian Liu}, \bibinfo{person}{Ke Wang},
  \bibinfo{person}{Yuchuan Wu}, \bibinfo{person}{Fei Huang},
  \bibinfo{person}{Yongbin Li}, \bibinfo{person}{Junge Zhang}, {and}
  \bibinfo{person}{Jianbin Jiao}.} \bibinfo{year}{2025}\natexlab{}.
\newblock \showarticletitle{Agentic Reinforcement Learning with Implicit Step
  Rewards}.
\newblock \bibinfo{journal}{\emph{arXiv preprint arXiv:2509.19199}}
  (\bibinfo{year}{2025}).
\newblock


\bibitem[{OpenAI}(2024)]%
        {openai2024gpt4osystemcard}
\bibfield{author}{\bibinfo{person}{{OpenAI}}.} \bibinfo{year}{2024}\natexlab{}.
\newblock \bibinfo{title}{GPT-4o System Card}.
\newblock
\showeprint[arxiv]{2410.21276}~[cs.CL]
\urldef\tempurl%
\url{https://arxiv.org/abs/2410.21276}
\showURL{%
\tempurl}


\bibitem[Park et~al\mbox{.}(2025)]%
        {park2025dipr1}
\bibfield{author}{\bibinfo{person}{Sungjune Park}, \bibinfo{person}{Hyunjun
  Kim}, \bibinfo{person}{Junho Kim}, \bibinfo{person}{Seongho Kim}, {and}
  \bibinfo{person}{Yong~Man Ro}.} \bibinfo{year}{2025}\natexlab{}.
\newblock \showarticletitle{DIP-R1: Deep Inspection and Perception with RL
  Looking Through and Understanding Complex Scenes}.
\newblock \bibinfo{journal}{\emph{arXiv preprint arXiv:2505.23179}}
  (\bibinfo{year}{2025}).
\newblock
\urldef\tempurl%
\url{https://arxiv.org/abs/2505.23179}
\showURL{%
\tempurl}


\bibitem[Ravi et~al\mbox{.}(2024)]%
        {SAM-2}
\bibfield{author}{\bibinfo{person}{Nikhila Ravi}, \bibinfo{person}{Valentin
  Gabeur}, \bibinfo{person}{Yuan-Ting Hu}, \bibinfo{person}{Ronghang Hu},
  \bibinfo{person}{Chaitanya Ryali}, \bibinfo{person}{Tengyu Ma},
  \bibinfo{person}{Haitham Khedr}, \bibinfo{person}{Roman Rädle},
  \bibinfo{person}{Chloe Rolland}, \bibinfo{person}{Laura Gustafson},
  \bibinfo{person}{Eric Mintun}, \bibinfo{person}{Junting Pan},
  \bibinfo{person}{Kalyan~Vasudev Alwala}, \bibinfo{person}{Nicolas Carion},
  \bibinfo{person}{Chao-Yuan Wu}, \bibinfo{person}{Ross Girshick},
  \bibinfo{person}{Piotr Dollár}, {and} \bibinfo{person}{Christoph
  Feichtenhofer}.} \bibinfo{year}{2024}\natexlab{}.
\newblock \bibinfo{title}{SAM 2: Segment Anything in Images and Videos}.
\newblock
\showeprint[arxiv]{2408.00714}~[cs.CV]
\urldef\tempurl%
\url{https://arxiv.org/abs/2408.00714}
\showURL{%
\tempurl}


\bibitem[Sarch et~al\mbox{.}(2025)]%
        {sarch2025grounded}
\bibfield{author}{\bibinfo{person}{Gabriel Sarch}, \bibinfo{person}{Snigdha
  Saha}, \bibinfo{person}{Naitik Khandelwal}, \bibinfo{person}{Ayush Jain},
  \bibinfo{person}{Michael~J. Tarr}, \bibinfo{person}{Aviral Kumar}, {and}
  \bibinfo{person}{Katerina Fragkiadaki}.} \bibinfo{year}{2025}\natexlab{}.
\newblock \showarticletitle{Grounded Reinforcement Learning for Visual
  Reasoning}. In \bibinfo{booktitle}{\emph{Advances in Neural Information
  Processing Systems (NeurIPS)}}.
\newblock
\urldef\tempurl%
\url{https://openreview.net/forum?id=1amnhVRQ3l}
\showURL{%
\tempurl}


\bibitem[Su et~al\mbox{.}(2025)]%
        {thinkingimagesmultimodalreasoning}
\bibfield{author}{\bibinfo{person}{Zhaochen Su}, \bibinfo{person}{Peng Xia},
  \bibinfo{person}{Hangyu Guo}, \bibinfo{person}{Zhenhua Liu},
  \bibinfo{person}{Yan Ma}, \bibinfo{person}{Xiaoye Qu}, \bibinfo{person}{Jiaqi
  Liu}, \bibinfo{person}{Yanshu Li}, \bibinfo{person}{Kaide Zeng},
  \bibinfo{person}{Zhengyuan Yang}, \bibinfo{person}{Linjie Li},
  \bibinfo{person}{Yu Cheng}, \bibinfo{person}{Heng Ji},
  \bibinfo{person}{Junxian He}, {and} \bibinfo{person}{Yi~R. Fung}.}
  \bibinfo{year}{2025}\natexlab{}.
\newblock \bibinfo{title}{Thinking with Images for Multimodal Reasoning:
  Foundations, Methods, and Future Frontiers}.
\newblock
\showeprint[arxiv]{2506.23918}~[cs.CV]
\urldef\tempurl%
\url{https://arxiv.org/abs/2506.23918}
\showURL{%
\tempurl}


\bibitem[Wang et~al\mbox{.}(2025d)]%
        {pixelreason}
\bibfield{author}{\bibinfo{person}{Haozhe Wang}, \bibinfo{person}{Alex Su},
  \bibinfo{person}{Weiming Ren}, \bibinfo{person}{Fangzhen Lin}, {and}
  \bibinfo{person}{Wenhu Chen}.} \bibinfo{year}{2025}\natexlab{d}.
\newblock \bibinfo{title}{Pixel Reasoner: Incentivizing Pixel-Space Reasoning
  with Curiosity-Driven Reinforcement Learning}.
\newblock
\showeprint[arxiv]{2505.15966}~[cs.CV]
\urldef\tempurl%
\url{https://arxiv.org/abs/2505.15966}
\showURL{%
\tempurl}


\bibitem[Wang et~al\mbox{.}(2025c)]%
        {wang_monet_2025}
\bibfield{author}{\bibinfo{person}{Qixun Wang}, \bibinfo{person}{Yang Shi},
  \bibinfo{person}{Yifei Wang}, \bibinfo{person}{Yuanxing Zhang},
  \bibinfo{person}{Pengfei Wan}, \bibinfo{person}{Kun Gai},
  \bibinfo{person}{Xianghua Ying}, {and} \bibinfo{person}{Yisen Wang}.}
  \bibinfo{year}{2025}\natexlab{c}.
\newblock \showarticletitle{Monet: {Reasoning} in {Latent} {Visual} {Space}
  {Beyond} {Images} and {Language}}.
\newblock  (\bibinfo{date}{Nov.} \bibinfo{year}{2025}).
\newblock
\href{https://doi.org/10.48550/arXiv.2511.21395}{doi:\nolinkurl{10.48550/arXiv.2511.21395}}
\newblock
\shownote{arXiv:2511.21395 [cs]}.


\bibitem[Wang et~al\mbox{.}(2025a)]%
        {wang2025divide}
\bibfield{author}{\bibinfo{person}{Wenbin Wang}, \bibinfo{person}{Liang Ding},
  \bibinfo{person}{Minyan Zeng}, \bibinfo{person}{Xiabin Zhou},
  \bibinfo{person}{Li Shen}, \bibinfo{person}{Yong Luo}, \bibinfo{person}{Wei
  Yu}, {and} \bibinfo{person}{Dacheng Tao}.} \bibinfo{year}{2025}\natexlab{a}.
\newblock \showarticletitle{Divide, conquer and combine: A training-free
  framework for high-resolution image perception in multimodal large language
  models}. In \bibinfo{booktitle}{\emph{Proceedings of the AAAI Conference on
  Artificial Intelligence}}, Vol.~\bibinfo{volume}{39}.
  \bibinfo{pages}{7907--7915}.
\newblock


\bibitem[Wang et~al\mbox{.}(2025b)]%
        {internVL3.5}
\bibfield{author}{\bibinfo{person}{Weiyun Wang}, \bibinfo{person}{Zhangwei
  Gao}, \bibinfo{person}{Lixin Gu}, \bibinfo{person}{Hengjun Pu},
  \bibinfo{person}{Long Cui}, \bibinfo{person}{Xingguang Wei},
  \bibinfo{person}{Zhaoyang Liu}, \bibinfo{person}{Linglin Jing},
  \bibinfo{person}{Shenglong Ye}, \bibinfo{person}{Jie Shao},
  \bibinfo{person}{Zhaokai Wang}, \bibinfo{person}{Zhe Chen},
  \bibinfo{person}{Hongjie Zhang}, \bibinfo{person}{Ganlin Yang},
  \bibinfo{person}{Haomin Wang}, \bibinfo{person}{Qi Wei},
  \bibinfo{person}{Jinhui Yin}, \bibinfo{person}{Wenhao Li},
  \bibinfo{person}{Erfei Cui}, \bibinfo{person}{Guanzhou Chen},
  \bibinfo{person}{Zichen Ding}, \bibinfo{person}{Changyao Tian},
  \bibinfo{person}{Zhenyu Wu}, \bibinfo{person}{Jingjing Xie},
  \bibinfo{person}{Zehao Li}, \bibinfo{person}{Bowen Yang},
  \bibinfo{person}{Yuchen Duan}, \bibinfo{person}{Xuehui Wang},
  \bibinfo{person}{Zhi Hou}, \bibinfo{person}{Haoran Hao},
  \bibinfo{person}{Tianyi Zhang}, \bibinfo{person}{Songze Li},
  \bibinfo{person}{Xiangyu Zhao}, \bibinfo{person}{Haodong Duan},
  \bibinfo{person}{Nianchen Deng}, \bibinfo{person}{Bin Fu},
  \bibinfo{person}{Yinan He}, \bibinfo{person}{Yi Wang},
  \bibinfo{person}{Conghui He}, \bibinfo{person}{Botian Shi},
  \bibinfo{person}{Junjun He}, \bibinfo{person}{Yingtong Xiong},
  \bibinfo{person}{Han Lv}, \bibinfo{person}{Lijun Wu}, \bibinfo{person}{Wenqi
  Shao}, \bibinfo{person}{Kaipeng Zhang}, \bibinfo{person}{Huipeng Deng},
  \bibinfo{person}{Biqing Qi}, \bibinfo{person}{Jiaye Ge},
  \bibinfo{person}{Qipeng Guo}, \bibinfo{person}{Wenwei Zhang},
  \bibinfo{person}{Songyang Zhang}, \bibinfo{person}{Maosong Cao},
  \bibinfo{person}{Junyao Lin}, \bibinfo{person}{Kexian Tang},
  \bibinfo{person}{Jianfei Gao}, \bibinfo{person}{Haian Huang},
  \bibinfo{person}{Yuzhe Gu}, \bibinfo{person}{Chengqi Lyu},
  \bibinfo{person}{Huanze Tang}, \bibinfo{person}{Rui Wang},
  \bibinfo{person}{Haijun Lv}, \bibinfo{person}{Wanli Ouyang},
  \bibinfo{person}{Limin Wang}, \bibinfo{person}{Min Dou},
  \bibinfo{person}{Xizhou Zhu}, \bibinfo{person}{Tong Lu},
  \bibinfo{person}{Dahua Lin}, \bibinfo{person}{Jifeng Dai},
  \bibinfo{person}{Weijie Su}, \bibinfo{person}{Bowen Zhou},
  \bibinfo{person}{Kai Chen}, \bibinfo{person}{Yu Qiao},
  \bibinfo{person}{Wenhai Wang}, {and} \bibinfo{person}{Gen Luo}.}
  \bibinfo{year}{2025}\natexlab{b}.
\newblock \bibinfo{title}{InternVL3.5: Advancing Open-Source Multimodal Models
  in Versatility, Reasoning, and Efficiency}.
\newblock
\showeprint[arxiv]{2508.18265}~[cs.CV]
\urldef\tempurl%
\url{https://arxiv.org/abs/2508.18265}
\showURL{%
\tempurl}


\bibitem[Wang et~al\mbox{.}(2025e)]%
        {thinklitevl}
\bibfield{author}{\bibinfo{person}{Xiyao Wang}, \bibinfo{person}{Zhengyuan
  Yang}, \bibinfo{person}{Chao Feng}, \bibinfo{person}{Hongjin Lu},
  \bibinfo{person}{Linjie Li}, \bibinfo{person}{Chung-Ching Lin},
  \bibinfo{person}{Kevin Lin}, \bibinfo{person}{Furong Huang}, {and}
  \bibinfo{person}{Lijuan Wang}.} \bibinfo{year}{2025}\natexlab{e}.
\newblock \showarticletitle{Sota with less: Mcts-guided sample selection for
  data-efficient visual reasoning self-improvement}.
\newblock \bibinfo{journal}{\emph{arXiv preprint arXiv:2504.07934}}
  (\bibinfo{year}{2025}).
\newblock


\bibitem[Wu and Xie(2024)]%
        {wu2024v}
\bibfield{author}{\bibinfo{person}{Penghao Wu} {and} \bibinfo{person}{Saining
  Xie}.} \bibinfo{year}{2024}\natexlab{}.
\newblock \showarticletitle{V?: Guided visual search as a core mechanism in
  multimodal llms}. In \bibinfo{booktitle}{\emph{Proceedings of the IEEE/CVF
  Conference on Computer Vision and Pattern Recognition}}.
  \bibinfo{pages}{13084--13094}.
\newblock


\bibitem[Yang et~al\mbox{.}(2023)]%
        {mmReact1}
\bibfield{author}{\bibinfo{person}{Zhengyuan Yang}, \bibinfo{person}{Linjie
  Li}, \bibinfo{person}{Jianfeng Wang}, \bibinfo{person}{Kevin Lin},
  \bibinfo{person}{Ehsan Azarnasab}, \bibinfo{person}{Faisal Ahmed},
  \bibinfo{person}{Zicheng Liu}, \bibinfo{person}{Ce Liu},
  \bibinfo{person}{Michael Zeng}, {and} \bibinfo{person}{Lijuan Wang}.}
  \bibinfo{year}{2023}\natexlab{}.
\newblock \bibinfo{title}{MM-REACT: Prompting ChatGPT for Multimodal Reasoning
  and Action}.
\newblock
\showeprint[arxiv]{2303.11381}~[cs.CV]
\urldef\tempurl%
\url{https://arxiv.org/abs/2303.11381}
\showURL{%
\tempurl}


\bibitem[Yao et~al\mbox{.}(2026)]%
        {yao2026prl}
\bibfield{author}{\bibinfo{person}{Jiarui Yao}, \bibinfo{person}{Ruida Wang},
  {and} \bibinfo{person}{Tong Zhang}.} \bibinfo{year}{2026}\natexlab{}.
\newblock \showarticletitle{PRL: Process Reward Learning Improves LLMs'
  Reasoning Ability and Broadens the Reasoning Boundary}.
\newblock \bibinfo{journal}{\emph{arXiv preprint arXiv:2601.10201}}
  (\bibinfo{year}{2026}).
\newblock


\bibitem[You and Wu(2025)]%
        {Seg-R1}
\bibfield{author}{\bibinfo{person}{Zuyao You} {and} \bibinfo{person}{Zuxuan
  Wu}.} \bibinfo{year}{2025}\natexlab{}.
\newblock \bibinfo{title}{Seg-R1: Segmentation Can Be Surprisingly Simple with
  Reinforcement Learning}.
\newblock
\showeprint[arxiv]{2506.22624}~[cs.CV]
\urldef\tempurl%
\url{https://arxiv.org/abs/2506.22624}
\showURL{%
\tempurl}


\bibitem[Yu et~al\mbox{.}(2025)]%
        {zoomrf1}
\bibfield{author}{\bibinfo{person}{Xuan Yu}, \bibinfo{person}{Dayan Guan},
  {and} \bibinfo{person}{Yanfeng Gu}.} \bibinfo{year}{2025}\natexlab{}.
\newblock \bibinfo{title}{Zoom-Refine: Boosting High-Resolution Multimodal
  Understanding via Localized Zoom and Self-Refinement}.
\newblock
\showeprint[arxiv]{2506.01663}~[cs.CV]
\urldef\tempurl%
\url{https://arxiv.org/abs/2506.01663}
\showURL{%
\tempurl}


\bibitem[Zhang et~al\mbox{.}(2025)]%
        {zhang_thyme_2025}
\bibfield{author}{\bibinfo{person}{Yi-Fan Zhang}, \bibinfo{person}{Xingyu Lu},
  \bibinfo{person}{Shukang Yin}, \bibinfo{person}{Chaoyou Fu},
  \bibinfo{person}{Wei Chen}, \bibinfo{person}{Xiao Hu}, \bibinfo{person}{Bin
  Wen}, \bibinfo{person}{Kaiyu Jiang}, \bibinfo{person}{Changyi Liu},
  \bibinfo{person}{Tianke Zhang}, \bibinfo{person}{Haonan Fan},
  \bibinfo{person}{Kaibing Chen}, \bibinfo{person}{Jiankang Chen},
  \bibinfo{person}{Haojie Ding}, \bibinfo{person}{Kaiyu Tang},
  \bibinfo{person}{Zhang Zhang}, \bibinfo{person}{Liang Wang},
  \bibinfo{person}{Fan Yang}, \bibinfo{person}{Tingting Gao}, {and}
  \bibinfo{person}{Guorui Zhou}.} \bibinfo{year}{2025}\natexlab{}.
\newblock \showarticletitle{Thyme: {Think} {Beyond} {Images}}.
\newblock  (\bibinfo{date}{Aug.} \bibinfo{year}{2025}).
\newblock
\href{https://doi.org/10.48550/arXiv.2508.11630}{doi:\nolinkurl{10.48550/arXiv.2508.11630}}
\newblock
\shownote{arXiv:2508.11630 [cs]}.


\bibitem[Zheng et~al\mbox{.}(2025)]%
        {zheng_deepeyes_2025}
\bibfield{author}{\bibinfo{person}{Ziwei Zheng}, \bibinfo{person}{Michael
  Yang}, \bibinfo{person}{Jack Hong}, \bibinfo{person}{Chenxiao Zhao},
  \bibinfo{person}{Guohai Xu}, \bibinfo{person}{Le Yang}, \bibinfo{person}{Chao
  Shen}, {and} \bibinfo{person}{Xing Yu}.} \bibinfo{year}{2025}\natexlab{}.
\newblock \showarticletitle{{DeepEyes}: {Incentivizing} "{Thinking} with
  {Images}" via {Reinforcement} {Learning}}.
\newblock  (\bibinfo{date}{May} \bibinfo{year}{2025}).
\newblock
\href{https://doi.org/10.48550/arXiv.2505.14362}{doi:\nolinkurl{10.48550/arXiv.2505.14362}}
\newblock
\shownote{arXiv:2505.14362 [cs]}.


\end{thebibliography}



\end{document}